\pdfoutput=1
\documentclass[10pt,twocolumn,letterpaper]{article}
\providecommand{\PaperMode}{preprint}

\newif\ifreviewversion
\edef\PaperModeValue{\PaperMode}%
\def\PaperModeReview{review}
\def\PaperModePreprint{preprint}
\ifx\PaperModeValue\PaperModeReview
  \reviewversiontrue
\else\ifx\PaperModeValue\PaperModePreprint
  \reviewversionfalse
\else
  \errmessage{config.tex: \string\PaperMode\space must be review or preprint}
\fi\fi
\usepackage[T1]{fontenc}
\ifreviewversion
  \usepackage[review]{cvpr}
\else
  \usepackage[pagenumbers]{cvpr}
\fi

\newcommand{\papertitle}{GenDAG: Rivaling Proprietary Image Generators\\with Open Models and Agentic Planning}

\PassOptionsToPackage{pdfversion=1.7}{hyperref}
\ifdefined\pdfminorversion
\else
  
\fi
\usepackage{graphicx}
\usepackage{amsmath,amssymb}
\usepackage{booktabs}
\usepackage{placeins}
\usepackage{colortbl,tabularx,multirow}
\definecolor{cvprblue}{rgb}{0.21,0.49,0.74}
\usepackage[pagebackref,breaklinks,colorlinks,allcolors=cvprblue]{hyperref}
\ifreviewversion\hypersetup{pdfauthor={}}\fi

\newcommand{\method}{GenDAG}

\definecolor{oursrow}{rgb}{0.91,0.95,0.98}
\definecolor{oursink}{rgb}{0.15,0.31,0.46}
\definecolor{tablegroup}{gray}{0.95}
\newcommand{\ourslabel}[1]{\textcolor{oursink}{\textbf{#1}}}

\renewcommand{\dbltopfraction}{0.8}
\title{\papertitle}
\ifreviewversion\else\newcommand{\affiliationmark}[1]{%
  \textsuperscript{\normalfont\color{black}\NoHyper#1\endNoHyper}%
}
\author{%
Chenxu~Bai\affiliationmark{1,2,*}\quad Yixin~Yang\affiliationmark{1,2}\quad Beier~Xiao\affiliationmark{2}\quad Chengxuan~Zhu\affiliationmark{1,2}\quad Chun~Chen\affiliationmark{1}\\
Qibin~Yang\affiliationmark{1,2}\quad Rang~Li\affiliationmark{1,2}\quad Sinan~Du\affiliationmark{1}\quad Taoyu~Yang\affiliationmark{1,2}\quad Yuechen~Zhang\affiliationmark{1}\\
Yuzhe~Weng\affiliationmark{1,3}\quad Zhaojun~Huang\affiliationmark{1,2}\quad Boxin~Shi\affiliationmark{2,\textdagger}\quad Shuhuai~Ren\affiliationmark{1,\textdagger}\\[4pt]
{\normalsize \affiliationmark{1}LLM Core, Xiaomi\qquad \affiliationmark{2}Peking University}\\
{\normalsize \affiliationmark{3}University of Science and Technology of China}\\[2pt]
{\small\ttfamily chenxu.bai@stu.pku.edu.cn\qquad shiboxin@pku.edu.cn}%
}

\newcommand{\authornotes}{%
  \begingroup
  \renewcommand{\thefootnote}{\fnsymbol{footnote}}%
  \footnotetext[1]{Contribution done during internship at Xiaomi MiMo Team.}%
  \footnotetext[2]{Co-corresponding authors.}%
  \endgroup
}
\fi

\begin{document}
\raggedbottom
\twocolumn[{
\renewcommand\twocolumn[1][]{#1}
\maketitle
\begin{minipage}{\textwidth}
\centering
\includegraphics[width=\linewidth]{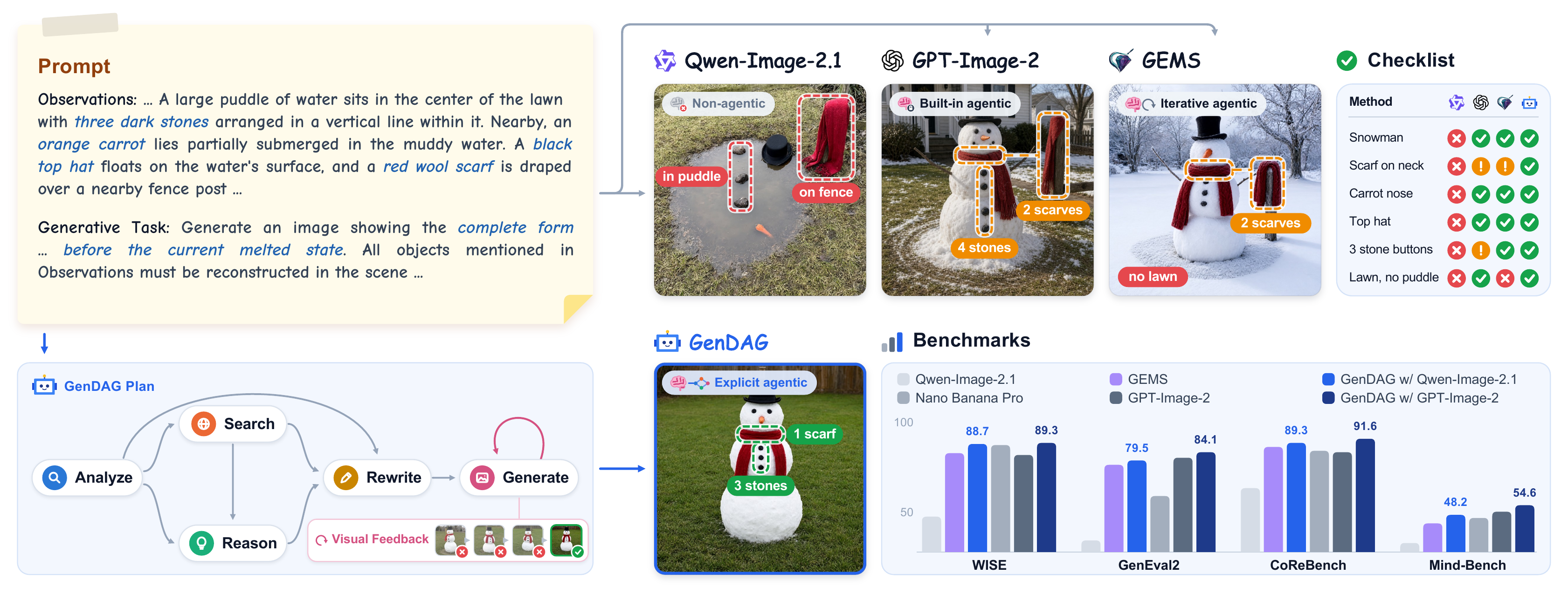}
\captionsetup{skip=4pt}
\captionof{figure}{\textbf{Agentic image generation with \method.} \textbf{(Top)} The prompt describes melted remnants on a lawn and asks for the scene before melting. Models must infer the unstated snowman while keeping the lawn, three stones, and one scarf. Qwen-Image-2.1 reproduces the melted scene. GPT-Image-2 and GEMS reconstruct the snowman but leave a second scarf on the fence. GPT-Image-2 adds a fourth stone, and GEMS covers the lawn with snow. \textbf{(Bottom)} \method{} plans analysis, retrieval, reasoning, generation, and refinement to satisfy all constraints. Our agent thus improves open generation and performs comparably to commercial GPT-Image-2 and Nano Banana Pro.}
\label{fig:teaser}
\end{minipage}
\par\vspace{8pt}
}]
\ifreviewversion\else\authornotes\fi
\ifreviewversion\begin{samepage}\fi
\begin{abstract}
Open text-to-image models struggle with requests that leave key content implicit or impose many explicit constraints. Multimodal agents can infer or retrieve this content and verify generated images, but for each request they must decide which operations to run and what information each one receives. We introduce \textbf{GenDAG}, an agentic framework that makes both decisions jointly by synthesizing a query-adaptive directed acyclic graph (DAG) whose edges carry typed intermediate results between operators. Before execution, a checker type-checks every edge and returns diagnostics for repair. After generation, visual checks against stated and inferred requirements guide regeneration. These mechanisms adapt each workflow reliably before execution and ground generation and verification in the same context to catch omissions and errors that detailed prompts miss. Using Qwen-Image-2.1 without fine-tuning, GenDAG scores 88.70 on WISE and 89.32 on T2I-CoReBench, exceeding direct generation with GPT-Image-2 and Nano Banana Pro, while trailing GPT-Image-2 by 1.51 points on GenEval2. Applied to GPT-Image-2, the same agent further improves the generation quality of this strong commercial model. Our project page is: \href{https://a1henu.github.io/GenDAG}{\upshape\small\ttfamily https://a1henu.github.io/GenDAG}

\end{abstract}
\ifreviewversion\end{samepage}\fi

\section{Introduction}
\label{sec:introduction}
Text-to-image (T2I) models have made substantial progress in visual quality, stylistic diversity, and prompt following~\cite{rombach2022ldm,podell2023sdxl,esser2024sd3,wu2025qwenimage}. However, complex requests that require inferring implicit content from world knowledge~\cite{li2025core,he2026mindbrush_sa,niu2026wise} and satisfying many explicit constraints remain challenging, particularly for open image generators. The reconstruction task in Fig.~\ref{fig:teaser} illustrates both difficulties. Its request describes remnants such as three stones in a puddle and a scarf on a fence post and asks for the scene before melting, but never names a snowman. An open image generator (e.g., Qwen-Image-2.1) reproduces these remnants and fails to infer the snowman they imply. A commercial model (e.g., GPT-Image-2) infers the snowman, likely through its built-in reasoning, but adds a fourth stone and leaves a second scarf on the fence. Both errors contradict the inferred scene, which has three stone buttons and one scarf on the snowman. These failures point to two limitations: First, open generators lack the reasoning needed to infer implicit content. Second, even though GPT-Image-2 and the GEMS agent~\cite{he2026gems} correctly infer the snowman, they still leave an extra scarf, which suggests that they do not use visual feedback to catch inconsistencies with the inferred scene.

Although open generators lack such reasoning, multimodal large language models (MLLMs) reason effectively over text and images~\cite{bai2025qwen3vl} and can judge whether an image satisfies stated requirements~\cite{hu2023tifa}. Agents built on large language models (LLMs) further plan multi-step solutions, call external tools such as web search, and revise outputs from feedback~\cite{yao2022react,gupta2022visprog,shinn2023reflexion}. These abilities let an MLLM agent address both limitations of image generators. Before generation, it infers or retrieves a request's implicit content and writes this content into a detailed generator prompt~\cite{he2026mindbrush_sa,zhang2026qia}. After generation, it checks the image against the stated and inferred requirements and corrects detected errors~\cite{wang2024genartist,he2026gems}.

Recent agentic image generators build on these capabilities through retrieval and iterative refinement~\cite{he2026mindbrush_sa,he2026gems}, adaptive tool use~\cite{ren2026scope}, workflow selection~\cite{chen2026genrouter}, and hierarchical planning~\cite{zhang2026qia}. Most execute predefined stages or choose fixed templates, so their design prescribes the information each operation receives. GEMS~\cite{he2026gems}, for example, writes its generation prompt and verification questions separately from the request text. In Fig.~\ref{fig:teaser}, its prompt places one scarf on the snowman and a second on the fence, and one question asks whether a red scarf is draped over the post. The image passes this check with both scarves, although the request mentions only one scarf, which belongs on the reconstructed snowman. Catching this duplicate requires one inferred scene to guide both steps, but the fixed GEMS pipeline shares no such scene between them. Other requests need different operations and connections, such as retrieval before rewriting for an unfamiliar entity or a checklist that guides refinement for a scene with many constraints. Since no fixed pipeline provides all these connections, planning for varied requests raises a key question: \emph{How can an agent jointly determine which operations to perform and what information each of these operations should consume?}

To address this question, we introduce \textbf{\method}, an agentic framework for image \textbf{Gen}eration through query-adaptive \textbf{D}irected \textbf{A}cyclic \textbf{G}raph (DAG) synthesis. For each request, \method{} composes a task-specific DAG from a catalog of twelve operators derived from approximately 156K agent trajectories and workflows. These operators range from retrieval, reasoning, and visual verification to code-rendered layouts, style selection, and image editing, and they cover 86.68\% of the recorded actions. Existing agents implement only subsets of these operators, such as verification without retrieval~\cite{wang2024genartist,he2026gems} or retrieval without image checks~\cite{he2026mindbrush_sa}, so some miss unstated knowledge and others leave errors in the image. Beyond this broader catalog, our agent plans which results each operator consumes, so that retrieved and inferred content reaches both the generation prompt and the verification checklist. To catch incompatible connections in agent-written plans, we equip operators with typed interfaces that distinguish retrieved evidence, generation prompts, and verification checklists. A checker verifies these connections before execution, and visual feedback against the stated and inferred content guides refinement.

Experiments on five benchmarks covering world knowledge, reasoning, and composition show that \method{} substantially improves the underlying open generator and performs comparably to leading commercial generators. Its plans also vary across benchmarks, with retrieval for knowledge-intensive requests and checklist-guided refinement for compositional ones. Moreover, typed diagnostics reduce the checker submissions needed to repair invalid plans. These results support planning explicit dependencies for each request and repairing plans with typed diagnostics.

Overall, our contributions are summarized as follows:
\begin{enumerate}
\item We build an operator catalog from approximately 156K agent trajectories and workflows and introduce typed interfaces, a checker for incompatible connections, and actionable diagnostics for repairing invalid plans.
\item We propose a query-adaptive DAG synthesis mechanism that jointly determines the operations required by each request and their information dependencies to coordinate reasoning, generation, and visual verification.
\item We demonstrate that the resulting agentic system can substantially improve open image generators and make them comparable to leading commercial generators.
\end{enumerate}

\begin{figure*}[t]
\centering
\includegraphics[width=\linewidth]{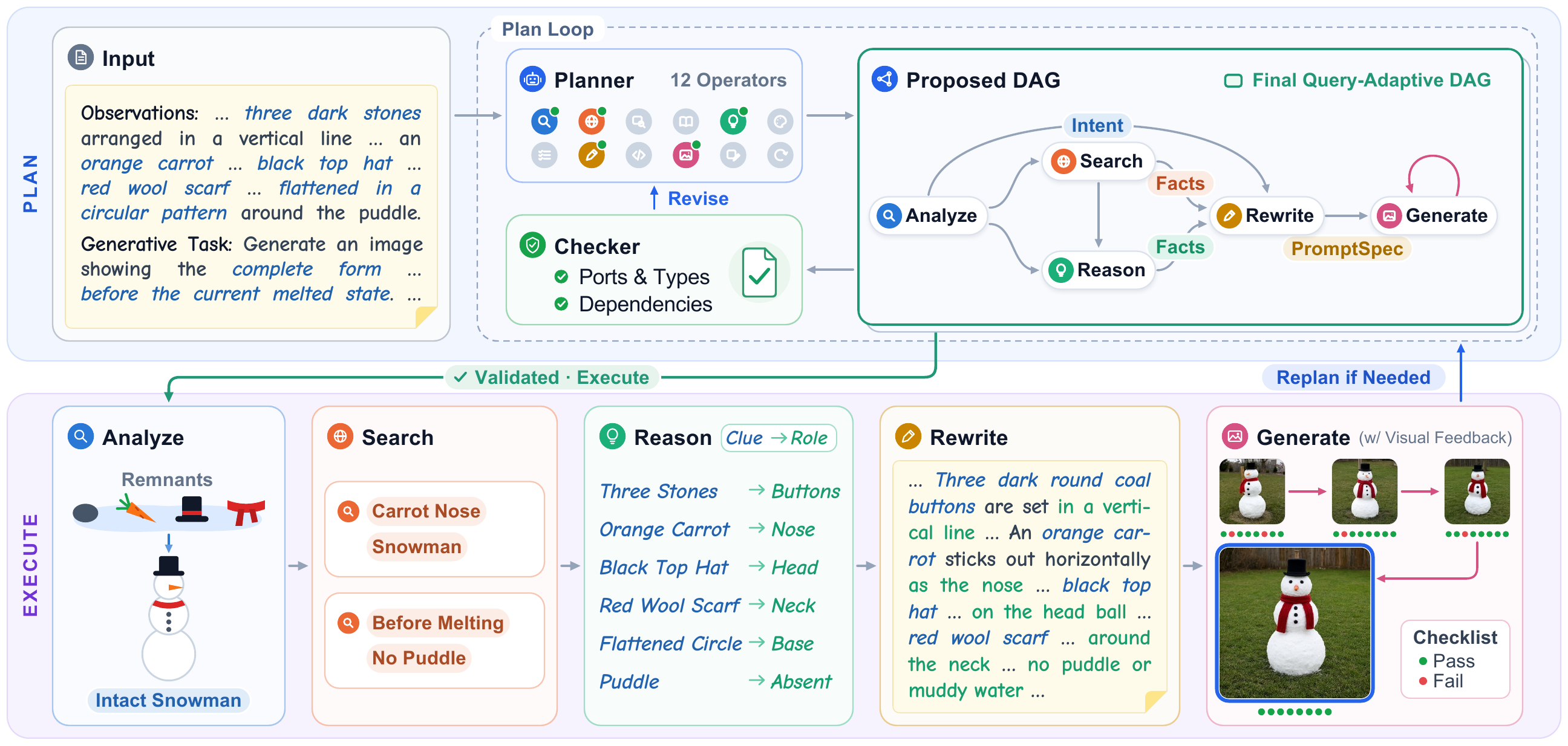}
\caption{\textbf{GenDAG overview.} The agent plans operators and their dependencies and repairs the DAG using diagnostics from a checker. The executor then passes retrieved and inferred context to generation, and the agent refines the image with visual feedback.}
\label{fig:overview}
\vspace{-4pt}
\end{figure*}
\section{Related Work}
\label{sec:related_work}
\paragraph{Agentic image generation.}
Early systems coordinate generation and editing tools, refine prompts through visual feedback, or organize multi-agent collaboration~\cite{wang2024genartist,yang2024idea2img,chen2025t2icopilot}. Reasoning-guided methods plan visual content before synthesis~\cite{fang2025got,jiang2025t2ir1}. Mind-Brush combines such reasoning with multimodal retrieval~\cite{he2026mindbrush_sa}, and SearchGen studies how learned and retrieved knowledge complement each other~\cite{wang2026searchgen}. GenClaw guides synthesis with code-generated sketches~\cite{ye2026genclaw}, and GEMS pairs verification with trajectory memory~\cite{he2026gems}. Further work explores multi-agent editing~\cite{yang2026m3}, accumulated experience~\cite{chen2026memogen}, and verifier-guided refinement~\cite{zhou2026stride}. Each method organizes its capabilities within a predefined pipeline, whereas \method{} composes operators from a shared catalog for each request.

\vspace{-3pt}
\paragraph{Adaptive planning for image generation.}
A system must also decide which capabilities each request needs and how results pass between steps. SCOPE maintains a semantic specification of entities and constraints to guide retrieval, reasoning, and repair~\cite{ren2026scope}. GenRouter selects workflow--generator pairs based on task requirements and expected quality, cost, and latency~\cite{chen2026genrouter}. Qwen-Image-Agent plans at the information, content, and generation levels, routes each missing piece of context to reasoning or search, and organizes these steps as dependency graphs for parallel execution~\cite{zhang2026qia}. \method{} instead jointly determines which operators to instantiate and what each one consumes. Unlike routing each missing piece of context to one strategy, joint planning lets operators build on each other, as when reasoning derives the intended scene from retrieved facts.

\vspace{-3pt}
\paragraph{Structured execution and plan validation.}
Explicit dependencies determine when operations can run and allow workflows to be checked before execution. ReAct interleaves reasoning and actions step by step~\cite{yao2022react}, while LLMCompiler organizes function calls as dependency graphs for parallel execution~\cite{kim2023llmcompiler}. GraSP composes agent skills into typed graphs with precondition--effect constraints, node-level verification, and local repair~\cite{xia2026grasp}, and ComfyClaw edits ComfyUI workflows as typed graphs with visual verification~\cite{li2026comfyclaw}. Their types describe skill preconditions and pipeline nodes, respectively. \method{} instead declares semantic types for operator inputs and outputs, such as retrieved evidence, allowing a checker to reject incompatible connections and guide targeted repairs before execution.

\section{Method}
\label{sec:method}

\subsection{Agent Overview}
\label{sec:agent_overview}
Given a text request $q$ and optional reference images $R$, \method{} produces an image with a frozen generator (Fig.~\ref{fig:overview}).
An MLLM serves as the agent's cognitive model for planning, reasoning, and verification.
For each request, the agent writes a plan (Sec.~\ref{sec:graph_synthesis}), a checker validates it before execution (Sec.~\ref{sec:plan_check}), and visual feedback corrects the resulting image (Sec.~\ref{sec:execution}).
We define twelve typed operators in Sec.~\ref{sec:operator_catalog}, and a plan composes them into a DAG $G=(V,E)$ whose nodes $v\in V$ are operator instances.
An edge $(u,v,p)\in E$ sends the output of node $u$ to the input port $p$, a named argument, of node $v$, so the edges state exactly which information each operator receives.
The request and reference images enter as external inputs, and a designated output node returns the image.
A plan thus fixes both the operators and the information that each one consumes.

\subsection{Typed Operators and Information Flow}
\label{sec:operator_catalog}
Existing image agents use fixed toolsets chosen by their designers: GEMS~\cite{he2026gems} verifies and refines images but cannot search the web, and Mind-Brush~\cite{he2026mindbrush_sa} retrieves knowledge but never inspects the image it produces.
A request that needs a missing operation cannot be served, however well the agent plans.
To find the operations used in practice, we collect approximately 156K agent trajectories and workflows from eight sources and group their actions into shared types.
We then add the frequent types our initial catalog lacks, namely code execution, image editing, and style selection.
The resulting twelve operators cover 86.68\% of all recorded actions, and their sources, mappings, and signatures are detailed in the supplementary material (Sec.~\ref{app:catalog}).

Each operator declares named input ports, an output type, and parameters.
\texttt{Analyze} and \texttt{Decompose} interpret the request, and \texttt{SearchText}, \texttt{SearchImage}, and \texttt{SkillQuery} retrieve evidence, reference images, and instructions from a bank of generation skills.
\texttt{Reason}, \texttt{Rewrite}, \texttt{SelectStyle}, and \texttt{CodeExec} turn this context into new facts, a generation prompt, a style, and a code-rendered layout.
\texttt{Generate} and \texttt{Edit} produce the image, and \texttt{RefineLoop} regenerates it against a checklist.

In a hand-built pipeline, the designer specifies each connection in advance, so each output is routed to its intended input. An agent that wires operators itself may instead send a checklist into the generation prompt, which would then render the check questions as image content. We therefore give every intermediate result a type that names its role.
\texttt{SearchText} and \texttt{Reason} produce \texttt{Facts} that carry evidence and inferred content, \texttt{Rewrite} combines them with the request into a generation description called a \texttt{PromptSpec}, and \texttt{Decompose} produces a \texttt{ChecklistSpec}.
\texttt{RefineLoop} thus receives the content to render and the criteria to check as separate inputs, and the checker (Sec.~\ref{sec:plan_check}) rejects mismatched connections.

\subsection{Query-Adaptive DAG Synthesis}
\label{sec:graph_synthesis}

\begin{figure*}[t]
\centering
\includegraphics[width=\linewidth]{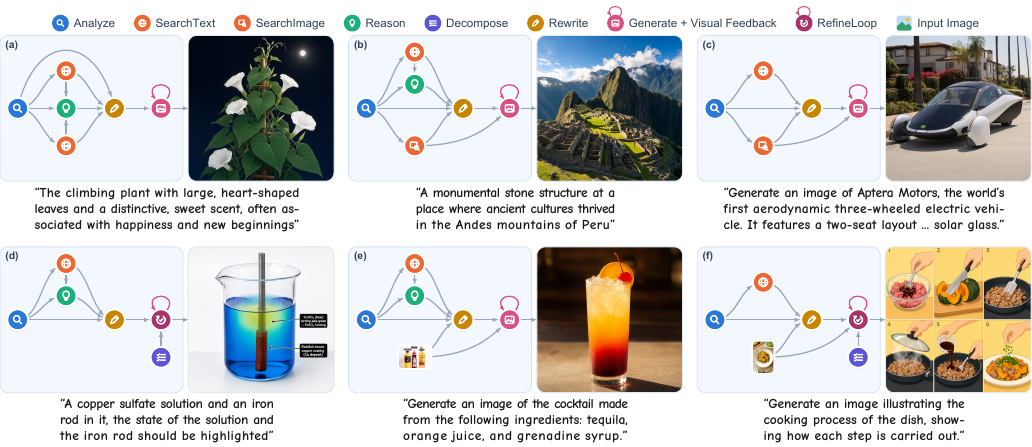}
\caption{\textbf{Query-adaptive DAGs from recorded executions.} Each panel shows a request (below), its planned DAG (left), and the Qwen-Image-2.1 output (right). \textbf{(a)}~Two searches retrieve the plant's identity and appearance, and \texttt{Reason} combines them. \textbf{(b)}~Facts and reference images are retrieved for the landmark, Machu Picchu. \textbf{(c)}~Facts and reference images are retrieved for the Aptera vehicle. \textbf{(d)}~\texttt{Reason} infers how iron reacts in copper sulfate, and \texttt{RefineLoop} checks the visible changes. \textbf{(e)}~The input image and listed ingredients identify the cocktail as a Tequila Sunrise. \textbf{(f)}~The recipe of the pictured dish is retrieved, and the step panels are refined against a checklist. Edges from the text request to every operator are omitted, and ellipses mark omitted prompt text.}
\label{fig:dag_cases}
\end{figure*}

Existing agents fix the flow of information in one of three ways.
Pipeline agents such as GEMS and Mind-Brush follow one fixed stage order, and GenRouter~\cite{chen2026genrouter} selects one of a few predefined workflow templates.
Step-by-step agents such as ReAct~\cite{yao2022react} choose the next tool after seeing each result.
The first two cannot add or reconnect steps for a request their designers did not anticipate, such as the snowman request, whose inferred scene must reach both the prompt and the checks.
Step-by-step agents can adapt, but they never form a complete plan, so their choices cannot be checked in advance and every step costs another model call.
\method{} instead writes the complete graph for each request before anything runs.
The agent first labels the request as \emph{complete}, \emph{incomplete}, or \emph{ambiguous} according to whether depicting it requires, or might require, knowledge that the request does not state.
It then lists the missing information and the requested constraints, picks the cheapest operator that can supply each missing piece, and connects every input to its source.
Requests with many checkable constraints, such as counts and rendered text, also receive a \texttt{Decompose} checklist that drives a \texttt{RefineLoop} node.
To write the graph, the agent starts from the most similar of ten small example plans and adds, removes, or reconnects nodes to fit the request.
The resulting plans differ across requests in both operators and connections (Fig.~\ref{fig:dag_cases}).

\subsection{Plan Checking and Diagnostic Repair}
\label{sec:plan_check}

A plan written for each request can contain mistakes that a hand-built pipeline never has, such as a missing node reference, an out-of-range parameter, a mistyped connection, or an empty required input.
Running such a plan wastes image generations or fails midway, and a bare error message does not say what to fix.
Before execution, a checker therefore verifies that the graph is acyclic and uses only known operators, that every node contributes to the output, and that every port receives inputs of the right type and number.
It also enforces parameter ranges, size limits such as at most 15 nodes and one \texttt{Generate} node, and two planning rules, one of which requires retrieval for incomplete or ambiguous requests.
The supplementary material lists all checks and rules (Sec.~\ref{app:checks}).
For an invalid plan, the checker returns a diagnostic that names the violated requirement, the affected node and port, and a repair hint such as the expected type.
The agent can therefore fix the faulty part and keep the rest of the plan, and after two failed repairs its instructions direct it to the closest example plan.
The instructions also forbid skipping the check or calling image backends directly, and the executor re-checks each plan before running it, so only plans that pass every check can produce images.

\subsection{Execution and Visual Feedback}
\label{sec:execution}
The executor runs the checked graph in dependency order, so retrieved and inferred content reaches the generator through the rewritten prompt and the wired reference images.
Generators still miss some requirements, and existing checks detect only some of these misses.
GEMS, for instance, checks its images with questions written from the request text, so it misses requirements that exist only in the inferred scene, such as the scarf belonging on the snowman.
\method{} therefore checks images at two levels.
Within the plan, \texttt{RefineLoop} checks each image against the \texttt{Decompose} checklist of explicit constraints and revises the prompt up to twice from the failed items.
Encapsulating this loop in one node keeps the plan acyclic.

After the plan runs, a final check also covers inferred content, with yes/no questions derived from the request, the acquired facts, and the final generation description.
Because the facts include inferred content, the questions can target the reconstructed snowman even though the request never names it.
For a question set $\mathcal{Q}=\{c_1,\ldots,c_m\}$, the agent scores a candidate image $I_t$, with $I_0$ denoting the output of the plan, by the number of checks that it passes:
\begin{equation}
s(I_t;\mathcal{Q})=\sum_{j=1}^{m}\mathbf{1}\bigl[\operatorname{VQA}(I_t,c_j)=\text{yes}\bigr].
\label{eq:visual_checks}
\end{equation}
Failed checks guide prompt revision and up to three regenerations without reference images, which stop once all checks pass.
The agent returns the candidate with the highest $s$ and prefers the earlier one in a tie, so a regeneration replaces the plan's output only when it passes more checks.
If a failed check reveals missing knowledge, the agent can revise the plan once by adding retrieval or reasoning.
These checks use the cognitive model, not benchmark judges.

\section{Experiments}
\label{sec:experiments}

\begin{table*}[t]
\begingroup
\renewenvironment{table*}[1][]{\ignorespaces}{\ignorespacesafterend}%
\begin{table*}[t]
\centering\footnotesize
\caption{\textbf{System comparison.} Scores are on a 0--100 scale with higher values indicating better performance. {\setlength{\fboxsep}{1pt}\colorbox{oursrow}{Blue rows}} identify GenDAG, and \textbf{bold} and \underline{underline} mark the highest and second-highest scores within each block. $^\dagger$ marks T2I-only Mind-Bench averages excluded from All-column ranking, and \textemdash\ denotes unsupported image inputs. Evaluation details are provided in Sec.~\ref{sec:setup}.}
\label{tab:main}
\setlength{\tabcolsep}{4.5pt}
\renewcommand{\arraystretch}{1.14}
\begin{tabular}{lrrrrrrr}
\toprule
\multirow{2}{*}[-\dimexpr(\aboverulesep+\belowrulesep+\cmidrulewidth)/2\relax]{\textbf{Method}} & \multicolumn{1}{c}{\multirow{2}{*}[-\dimexpr(\aboverulesep+\belowrulesep+\cmidrulewidth)/2\relax]{\textbf{WISE}}} & \multicolumn{1}{c}{\multirow{2}{*}[-\dimexpr(\aboverulesep+\belowrulesep+\cmidrulewidth)/2\relax]{\textbf{GenEval2}}} & \multicolumn{1}{c}{\multirow{2}{*}[-\dimexpr(\aboverulesep+\belowrulesep+\cmidrulewidth)/2\relax]{\textbf{T2I-CoReBench}}} & \multicolumn{1}{c}{\multirow{2}{*}[-\dimexpr(\aboverulesep+\belowrulesep+\cmidrulewidth)/2\relax]{\textbf{AgentGen}}} & \multicolumn{3}{c}{\textbf{Mind-Bench}} \\
\cmidrule(lr){6-8}
 & & & & & \multicolumn{1}{c}{\textbf{T2I}} & \multicolumn{1}{c}{\textbf{I2I}} & \multicolumn{1}{c}{\textbf{All}} \\
\midrule
\rowcolor{tablegroup}\multicolumn{8}{l}{\strut\textit{Commercial image generation backends}} \\
GPT-Image-2 & 82.60 & \underline{81.04} & 84.13 & \textbf{73.52} & \underline{59.71} & 30.67 & \underline{51.00} \\
Nano Banana Pro & \underline{88.10} & 59.73 & \underline{84.91} & 62.64 & 48.29 & \underline{34.67} & 44.20 \\
\rowcolor{oursrow}
\ourslabel{GenDAG w/ GPT-Image-2} & \textbf{89.30} & \textbf{84.12} & \textbf{91.60} & \underline{68.75} & \textbf{60.00} & \textbf{42.00} & \textbf{54.60} \\
\midrule
\rowcolor{tablegroup}\multicolumn{8}{l}{\strut\textit{Open-source generator: Qwen-Image-2.1}} \\
Qwen-Image-2.1 & 45.90 & 15.00 & 64.13 & 33.07 & 8.57 & 1.33 & 6.40 \\
Qwen-Image-2.1-PE & 71.30 & 55.19 & 73.40 & \textbf{59.28} & 28.29 & \underline{38.00} & 31.20 \\
GEMS$^\dagger$ & 83.60 & \underline{77.09} & \underline{87.06} & 56.60 & 37.14 & \textemdash & 37.14 \\
Mind-Brush & 84.40 & 50.86 & 82.86 & 44.38 & 50.86 & 26.00 & 43.40 \\
GenRouter$^\dagger$ & 83.30 & 67.79 & 80.16 & 44.32 & 46.57 & \textemdash & 46.57 \\
SCOPE & \underline{85.90} & 69.19 & 85.45 & 57.61 & \textbf{63.43} & 37.33 & \textbf{55.60} \\
\rowcolor{oursrow}
\ourslabel{GenDAG w/ Qwen-Image-2.1} & \textbf{88.70} & \textbf{79.53} & \textbf{89.32} & \underline{57.67} & \underline{52.00} & \textbf{39.33} & \underline{48.20} \\
\bottomrule
\end{tabular}
\end{table*}

\vspace{6pt}
\begin{table*}[t]
\centering\footnotesize
\caption{\textbf{T2I-CoReBench capability breakdown.} Composition covers multi-instance (MI), multi-attribute (MA), multi-relation (MR), and text rendering (TR). Reasoning covers logical (LR), reconstructive (RR), hypothetical (HR), behavioral (BR), generalization (GR), commonsense (CR), analogical (AR), and procedural reasoning (PR). $\Delta$ is the gain over direct generation. }
\label{tab:core_main}
\setlength{\tabcolsep}{1.8pt}
\renewcommand{\arraystretch}{1.14}
\begin{tabularx}{\textwidth}{l>{\hsize=1.000000\hsize\linewidth=\hsize\raggedleft\arraybackslash}X>{\hsize=1.000000\hsize\linewidth=\hsize\raggedleft\arraybackslash}X>{\hsize=1.000000\hsize\linewidth=\hsize\raggedleft\arraybackslash}X>{\hsize=1.000000\hsize\linewidth=\hsize\raggedleft\arraybackslash}X>{\hsize=1.000000\hsize\linewidth=\hsize\raggedleft\arraybackslash}X>{\hsize=1.000000\hsize\linewidth=\hsize\raggedleft\arraybackslash}X>{\hsize=1.000000\hsize\linewidth=\hsize\raggedleft\arraybackslash}X>{\hsize=1.000000\hsize\linewidth=\hsize\raggedleft\arraybackslash}X>{\hsize=1.000000\hsize\linewidth=\hsize\raggedleft\arraybackslash}X>{\hsize=1.000000\hsize\linewidth=\hsize\raggedleft\arraybackslash}X>{\hsize=1.000000\hsize\linewidth=\hsize\raggedleft\arraybackslash}X>{\hsize=1.000000\hsize\linewidth=\hsize\raggedleft\arraybackslash}X>{\hsize=1.000000\hsize\linewidth=\hsize\raggedleft\arraybackslash}X>{\hsize=1.000000\hsize\linewidth=\hsize\raggedleft\arraybackslash}X>{\hsize=1.000000\hsize\linewidth=\hsize\raggedleft\arraybackslash}X}
\toprule
 & \multicolumn{5}{c}{Composition} & \multicolumn{9}{c}{Reasoning} & \\
\cmidrule(lr){2-6}\cmidrule(lr){7-15}
\textbf{Method} & \textbf{MI} & \textbf{MA} & \textbf{MR} & \textbf{TR} & \textbf{C-Avg} & \textbf{LR} & \textbf{RR} & \textbf{HR} & \textbf{BR} & \textbf{GR} & \textbf{CR} & \textbf{AR} & \textbf{PR} & \textbf{R-Avg} & \textbf{All} \\
\midrule
\rowcolor{tablegroup}\multicolumn{16}{l}{\strut\textit{Commercial generators}} \\
GPT-Image-2 & \underline{96.79} & \underline{94.86} & \underline{88.55} & \textbf{99.39} & \underline{94.90} & 69.70 & 55.05 & 71.95 & 71.00 & \underline{92.23} & 80.18 & \underline{92.52} & \textbf{97.30} & 78.74 & 84.13 \\
Nano Banana Pro & 95.83 & 93.74 & 86.39 & 99.08 & 93.76 & \underline{93.23} & \underline{76.71} & \underline{79.07} & \underline{72.89} & 71.05 & \underline{84.37} & 87.06 & 79.45 & \underline{80.48} & \underline{84.91} \\
\rowcolor{oursrow}
\ourslabel{GenDAG w/ GPT-Image-2} & \textbf{97.27} & \textbf{95.14} & \textbf{91.48} & \underline{99.16} & \textbf{95.76} & \textbf{97.86} & \textbf{80.48} & \textbf{88.53} & \textbf{80.47} & \textbf{93.15} & \textbf{85.89} & \textbf{92.87} & \underline{96.89} & \textbf{89.52} & \textbf{91.60} \\
\rowcolor{oursrow}
\textcolor{oursink}{\textit{$\Delta$ vs.\ direct}} & \textcolor{oursink}{+0.48} & \textcolor{oursink}{+0.29} & \textcolor{oursink}{+2.93} & \textcolor{oursink}{-0.23} & \textcolor{oursink}{+0.87} & \textcolor{oursink}{+28.16} & \textcolor{oursink}{+25.43} & \textcolor{oursink}{+16.58} & \textcolor{oursink}{+9.47} & \textcolor{oursink}{+0.92} & \textcolor{oursink}{+5.71} & \textcolor{oursink}{+0.35} & \textcolor{oursink}{-0.42} & \textcolor{oursink}{+10.78} & \textcolor{oursink}{+7.47} \\
\midrule
\rowcolor{tablegroup}\multicolumn{16}{l}{\strut\textit{Open-source generator: Qwen-Image-2.1}} \\
Qwen-Image-2.1 & 89.57 & 90.68 & 72.98 & 94.88 & 87.03 & 43.06 & 18.02 & 54.68 & 34.23 & 61.28 & 59.57 & 65.26 & 85.39 & 52.69 & 64.13 \\
Qwen-Image-2.1-PE & 90.72 & 90.15 & 77.29 & 97.69 & 88.96 & 59.80 & 37.10 & 62.42 & 49.32 & 77.99 & 70.72 & 76.41 & 91.13 & 65.61 & 73.40 \\
GEMS & \textbf{95.71} & \underline{93.71} & \textbf{87.78} & \textbf{98.84} & \textbf{94.01} & 95.84 & 67.41 & \underline{78.04} & \underline{70.17} & \underline{91.51} & 81.04 & \textbf{88.75} & \underline{95.88} & \underline{83.58} & \underline{87.06} \\
Mind-Brush & 91.70 & 92.00 & 76.25 & 95.09 & 88.76 & 89.14 & \underline{72.56} & 69.28 & 66.92 & 87.27 & 78.12 & 83.94 & 92.05 & 79.91 & 82.86 \\
GenRouter & 92.42 & 91.59 & 77.26 & 95.95 & 89.31 & 62.88 & 68.36 & 70.33 & 63.70 & 86.28 & 77.31 & 83.52 & 92.34 & 75.59 & 80.16 \\
SCOPE & \underline{95.34} & 93.20 & 82.92 & 96.73 & 92.05 & \underline{96.33} & 62.71 & 76.96 & 63.51 & \textbf{92.05} & \underline{82.40} & \underline{88.43} & 94.86 & 82.16 & 85.45 \\
\rowcolor{oursrow}
\ourslabel{GenDAG w/ Qwen-Image-2.1} & 94.78 & \textbf{94.21} & \underline{87.00} & \underline{98.08} & \underline{93.52} & \textbf{97.22} & \textbf{79.39} & \textbf{84.07} & \textbf{76.84} & 91.44 & \textbf{84.13} & 88.08 & \textbf{96.63} & \textbf{87.23} & \textbf{89.32} \\
\rowcolor{oursrow}
\textcolor{oursink}{\textit{$\Delta$ vs.\ direct}} & \textcolor{oursink}{+5.21} & \textcolor{oursink}{+3.53} & \textcolor{oursink}{+14.02} & \textcolor{oursink}{+3.20} & \textcolor{oursink}{+6.49} & \textcolor{oursink}{+54.16} & \textcolor{oursink}{+61.37} & \textcolor{oursink}{+29.40} & \textcolor{oursink}{+42.62} & \textcolor{oursink}{+30.16} & \textcolor{oursink}{+24.56} & \textcolor{oursink}{+22.82} & \textcolor{oursink}{+11.24} & \textcolor{oursink}{+34.54} & \textcolor{oursink}{+25.19} \\
\bottomrule
\end{tabularx}
\end{table*}

\endgroup
\end{table*}

\subsection{Experimental Setup}
\label{sec:benchmarks}
\label{sec:setup}

\paragraph{Benchmarks and Metrics.}
We evaluate on WISE~\cite{niu2026wise} for world knowledge, GenEval2~\cite{kamath2025geneval2} for composition, T2I-CoReBench~\cite{li2025core} for composition and reasoning, AgentGen-Bench~\cite{wang2026searchgen} for knowledge-intensive generation with rendered text, and Mind-Bench~\cite{he2026mindbrush_sa} for knowledge and reasoning in generation and editing.
We score one final image per request, except on T2I-CoReBench, where we average four executions.
We use WISE's verified protocol with Qwen3.5-35B-A3B, GenEval2's soft TIFA geometric mean with Qwen3-VL-8B-Instruct, and T2I-CoReBench's Qwen3.5-9B scoring branch.
AgentGen-Bench averages its nine applicable criteria (Overall9), and Mind-Bench uses checklist satisfaction accuracy (CSA) with Gemini 3 Flash.
Scores use a 0--100 scale, and failures count as zero.

\paragraph{Baselines.}
On the same Qwen-Image-2.1 generator~\cite{teamqwen}, we compare direct generation and official prompt enhancement (PE) with four image agents: GEMS~\cite{he2026gems}, Mind-Brush~\cite{he2026mindbrush_sa}, GenRouter~\cite{chen2026genrouter}, and SCOPE~\cite{ren2026scope}.
GPT-Image-2 and Nano Banana Pro serve as commercial direct-generation references, and we pair \method{} with GPT-Image-2 to test agentic generation on a stronger backend.
All methods are scored on the requests they support by the same benchmark judges under the same protocols.

\paragraph{Implementation Details.}
All image agents use MiMo-V2.6-Flash~\cite{xiaomi2026mimov26} as the cognitive model.
In the Qwen-Image-2.1 group, all methods use 40 sampling steps, and the agents share a frozen retrieval environment.
We adapt the released agents but keep their original workflows and iteration limits, as detailed in the supplementary material (Sec.~\ref{app:baselines}).
Only GenRouter receives benchmark-score feedback, which it uses to initialize routing and update memory.

\begin{figure*}[t]
\centering
\includegraphics[width=\textwidth]{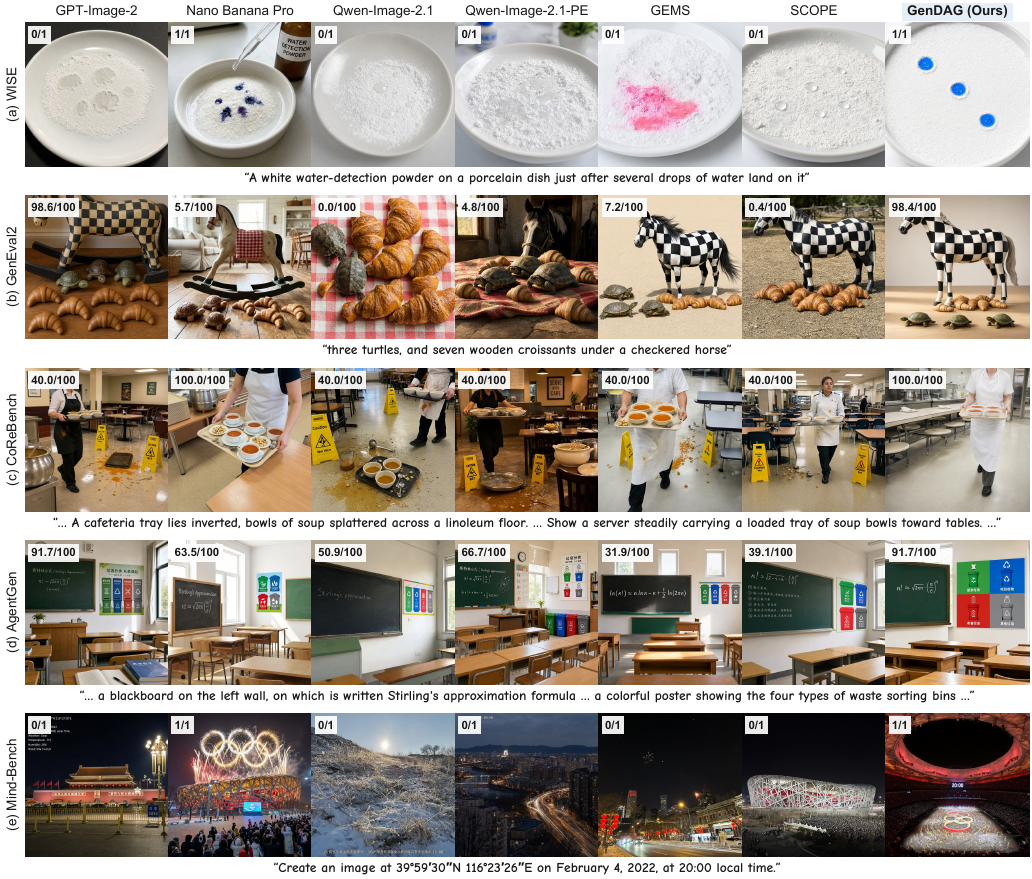}
\caption{\textbf{Qualitative comparison on matched prompts.} Each row presents one curated prompt per benchmark, and each column shows a method's first execution. On each request, our agent scores comparably to the higher-scoring commercial generator. Badges give score/maximum, where WiScore and CSA are binary, and soft TIFA (GM), CoRe score, and Overall9 are scaled to 100.}
\label{fig:qualitative}
\end{figure*}
\subsection{Main Results}
\label{sec:main_results}
\paragraph{Comparison with Open Agents.}

\method{} improves the same open generator across all five benchmarks (Table~\ref{tab:main}).
Over direct Qwen-Image-2.1, it gains 42.80 points on WISE, 64.54 on GenEval2, and 25.19 on T2I-CoReBench.
It also outperforms PE and every compared image agent on these three benchmarks, and its WISE gains cover all six knowledge categories (32.50 in physics to 57.50 in chemistry). Knowledge acquisition, reasoning, and visual feedback thus substantially strengthen the frozen generator.

\paragraph{Comparison with Commercial Generators.}

The resulting open system is competitive with the commercial references, as it exceeds both direct generators on WISE and T2I-CoReBench and trails GPT-Image-2 by 1.51 points on GenEval2.
On Mind-Bench, its overall score of 48.20 is only 2.80 points below GPT-Image-2 and above Nano Banana Pro (44.20).
Paired with GPT-Image-2, the same agent achieves the highest scores among all compared systems on WISE, GenEval2, T2I-CoReBench, and Mind-Bench I2I.

\paragraph{Capability Breakdown.}
\label{sec:capabilities}

On T2I-CoReBench (Table~\ref{tab:core_main}), \method{} raises the reasoning average over eight dimensions from 52.69 to 87.23 and the composition average over four dimensions from 87.03 to 93.52. Its reasoning average also exceeds that of GEMS (83.58), the strongest same-generator baseline.
Logical reasoning (43.06 to 97.22) and reconstructive reasoning (18.02 to 79.39) gain the most. These gains support inferring what a scene should depict before synthesis, as the snowman example requires.
On logical, reconstructive, hypothetical, and behavioral reasoning, \method{} also outperforms both commercial generators.

\vspace{-1pt}
\paragraph{Qualitative Results.}
\label{sec:qualitative}
In Fig.~\ref{fig:qualitative}, \method{} meets visible requirements that the displayed open baselines miss.
On WISE, the water-detection powder turns blue where water lands.
On T2I-CoReBench, the request lists the aftermath of a soup spill and asks for the earlier moment when a server carries the tray, and \method{} shows a clean floor without spill or warning signs.
It also identifies the Beijing Winter Olympics opening ceremony from coordinates and a timestamp in Mind-Bench and renders Stirling's approximation and a poster of four labeled waste bins in AgentGen-Bench.

\subsection{Generalization across Image Backends}
\label{sec:backends}
\label{sec:commercial_generator}
\begin{table}[t]
\centering\footnotesize
\caption{\textbf{Generalization across backends.} GenDAG consistently improves performance across diverse image generators.}
\label{tab:backend_generalization}
\setlength{\tabcolsep}{2.1pt}
\renewcommand{\arraystretch}{1.14}
\begin{tabular}{lrrrrr}
\toprule
\textbf{Method} & \textbf{WISE} & \textbf{GenEval2} & \textbf{CoReBench} & \textbf{AgentGen} & \textbf{Mind-Bench} \\
\midrule
\rowcolor{tablegroup}\multicolumn{6}{l}{\strut\textit{Backend: GPT-Image-2}} \\
Direct & 82.60 & 81.04 & 84.13 & 73.52 & 51.00 \\
\rowcolor{oursrow}
\ourslabel{GenDAG} & 89.30 & 84.12 & 91.60 & 68.75 & 54.60 \\
\rowcolor{oursrow}
\textcolor{oursink}{\textit{$\Delta$ vs.\ direct}} & \textcolor{oursink}{+6.70} & \textcolor{oursink}{+3.07} & \textcolor{oursink}{+7.47} & \textcolor{oursink}{-4.78} & \textcolor{oursink}{+3.60} \\
\midrule
\rowcolor{tablegroup}\multicolumn{6}{l}{\strut\textit{Backend: Qwen-Image-2512 / Edit-2511}} \\
Direct & 48.10 & 35.06 & 65.90 & 34.29 & 9.00 \\
\rowcolor{oursrow}
\ourslabel{GenDAG} & 86.20 & 75.41 & 86.96 & 49.51 & 39.00 \\
\rowcolor{oursrow}
\textcolor{oursink}{\textit{$\Delta$ vs.\ direct}} & \textcolor{oursink}{+38.10} & \textcolor{oursink}{+40.35} & \textcolor{oursink}{+21.05} & \textcolor{oursink}{+15.23} & \textcolor{oursink}{+30.00} \\
\midrule
\rowcolor{tablegroup}\multicolumn{6}{l}{\strut\textit{Backend: Qwen-Image-2.1}} \\
Direct & 45.90 & 15.00 & 64.13 & 33.07 & 6.40 \\
\rowcolor{oursrow}
\ourslabel{GenDAG} & 88.70 & 79.53 & 89.32 & 57.67 & 48.20 \\
\rowcolor{oursrow}
\textcolor{oursink}{\textit{$\Delta$ vs.\ direct}} & \textcolor{oursink}{+42.80} & \textcolor{oursink}{+64.54} & \textcolor{oursink}{+25.19} & \textcolor{oursink}{+24.60} & \textcolor{oursink}{+41.80} \\
\bottomrule
\end{tabular}
\end{table}

To test whether these gains depend on the generator, we compare direct and agent-assisted generation separately with GPT-Image-2 and two Qwen backends (Table~\ref{tab:backend_generalization}).
\method{} raises scores on WISE, GenEval2, T2I-CoReBench, and Mind-Bench with all three backends and on all five benchmarks with both Qwen versions.
With GPT-Image-2, its gains over direct generation still reach 7.47 points on T2I-CoReBench and 6.70 on WISE.
These gains show that the agent transfers across generators.

\subsection{Analysis of Plans and Visual Feedback}
\label{sec:analysis}
\label{sec:ablations}

\paragraph{Plans Across Benchmarks.}
\label{sec:plan_adaptivity}
The agent adapts its plans to what each benchmark leaves unspecified.
Across the 7,368 successful executions in Table~\ref{tab:main}, it produces 136 distinct operator graphs, and only 8.46\% of these plans reduce to the \texttt{Analyze}--\texttt{Rewrite}--\texttt{Generate} path.
Plans for knowledge requests draw on retrieval, with \texttt{SearchText} in 81.50\% of WISE plans but only 9.25\% of GenEval2 plans.
On compositional requests, plans instead use checklist refinement, with \texttt{RefineLoop} in 74.13\% of GenEval2 plans but only 2.80\% of WISE plans.
Plan statistics for every benchmark can be found in the supplementary material (Sec.~\ref{app:plan_stats}).

\vspace{-2pt}\paragraph{Visual Feedback.}
Even with detailed prompts, generated images often miss some requirements.
Only 17.75\% of the images returned by the planned graph pass all post-execution checks, which makes visual feedback essential.
A regeneration replaces the graph output in 67.05\% of executions, and 58.31\% of plans also use \texttt{RefineLoop}.

\subsection{DAG Planning versus Sequential Tool Use}
\label{sec:composition}
\begin{table}[t]
\centering\footnotesize
\caption{\textbf{DAG planning and sequential tool use.} LLM (planning and operator calls), Search, and Image (generation and editing) count requests per successful execution, including retries. Tokens likewise counts reported LLM tokens with cached input. Scores include failed runs, and \textbf{bold} marks the best per benchmark.}
\label{tab:structure}
\setlength{\tabcolsep}{3.8pt}\renewcommand{\arraystretch}{1.12}
\begin{tabular}{lrrrrr}
\toprule
\textbf{Policy} & \textbf{Score $\uparrow$} & \textbf{LLM $\downarrow$} & \textbf{Tokens (k) $\downarrow$} & \textbf{Search $\downarrow$} & \textbf{Image $\downarrow$} \\
\midrule
\rowcolor{tablegroup}\multicolumn{6}{l}{\strut\textit{WISE}} \\
ReAct & 88.30 & 61.13 & 564.35 & 1.55 & \textbf{2.96} \\
\rowcolor{oursrow}
\ourslabel{Ours} & \textbf{88.70} & \textbf{53.94} & \textbf{340.32} & \textbf{1.31} & 3.23 \\
\midrule
\rowcolor{tablegroup}\multicolumn{6}{l}{\strut\textit{T2I-CoReBench}} \\
ReAct & \textbf{89.62} & 79.69 & 704.03 & 0.39 & \textbf{4.52} \\
\rowcolor{oursrow}
\ourslabel{Ours} & 89.32 & \textbf{68.60} & \textbf{380.21} & \textbf{0.28} & 4.85 \\
\bottomrule
\end{tabular}
\end{table}

To compare explicit dependency planning with step-by-step tool use, we evaluate \method{} against a ReAct agent~\cite{yao2022react} that selects one operator at a time based on previous observations.
Both agents share models, operators, verification, and retrieval, and they differ only in how they plan and in their own step limits.
Details of both agents and counts with failed attempts can be found in the supplementary material (Sec.~\ref{app:react}).

\method{} and ReAct score similarly, with 88.70 versus 88.30 on WISE and 89.32 versus 89.62 on T2I-CoReBench (Table~\ref{tab:structure}).
\method{} reaches these scores with 11.76\% fewer cognitive-model requests per successful execution on WISE and 13.92\% fewer on T2I-CoReBench.
This reduction also holds when failed attempts are counted.
\method{} also needs fewer retrieval requests and only 60.30\% and 54.01\% of ReAct's reported LLM tokens on WISE and T2I-CoReBench, respectively.
\method{} specifies its acquisition and synthesis steps in one checked plan, whereas ReAct chooses each operator after observing the previous result and therefore queries its cognitive model at every step.

\subsection{Plan Repair with Typed Diagnostics}
\label{sec:reliability}
\begin{table}[t]
\centering\footnotesize
\caption{\textbf{Repair with typed checker diagnostics.} Fix@$k$ reports repairs (\%) within $k$ checker submissions. The Checks column averages checker submissions across all erroneous proposals, including unresolved ones, under a two-submission limit.}
\label{tab:reliability}
\setlength{\tabcolsep}{3.8pt}\renewcommand{\arraystretch}{1.12}
\begin{tabular}{lrrr}
\toprule
\textbf{Feedback} & \textbf{Fix@1 (\%)} & \textbf{Fix@2 (\%)} & \textbf{Checks $\downarrow$} \\
\midrule
\rowcolor{tablegroup}\multicolumn{4}{l}{\strut\textit{Backend: MiMo-V2.5 ($n=42$)}} \\
Rejection only & 69.05 & 95.24 & 1.31 \\
\rowcolor{oursrow}
Typed diagnostics & \textbf{92.86} & \textbf{100.00} & \textbf{1.07} \\
\midrule
\rowcolor{tablegroup}\multicolumn{4}{l}{\strut\textit{Backend: MiMo-V2.6-Flash ($n=48$)}} \\
Rejection only & 85.42 & 97.92 & 1.15 \\
\rowcolor{oursrow}
Typed diagnostics & \textbf{95.83} & \textbf{100.00} & \textbf{1.06} \\
\bottomrule
\end{tabular}
\end{table}

With two cognitive models, we test whether checker diagnostics help the agent repair invalid plans.
For each model, both conditions start from the same invalid first plans that the checker rejected in normal runs and differ only in whether the agent receives a bare rejection or typed diagnostics. On WISE and T2I-CoReBench, the cohorts contain 42 invalid first plans from MiMo-V2.5 planning-only sessions and all 48 from MiMo-V2.6-Flash executions. Each repair allows at most two checks and generates no images. The supplementary material details both cohorts (Sec.~\ref{app:repair}).

Diagnostics increase first-repair success from 69.05\% to 92.86\% with MiMo-V2.5 and from 85.42\% to 95.83\% with MiMo-V2.6-Flash (Table~\ref{tab:reliability}).
With diagnostics, the agent also repairs every proposal within two checks, and mean submissions fall from 1.31 to 1.07 and from 1.15 to 1.06.

\section{Conclusion}
\label{sec:conclusion}
We presented \method{}, an agentic image generation framework based on query-adaptive DAG synthesis.
For each request, the agent jointly selects the operators and the information each consumes, and a checker type-checks the plan before execution.
Without retraining, the framework substantially improves the open generator and makes it competitive with commercial generators on knowledge- and reasoning-intensive tasks, and the same agent also transfers to GPT-Image-2.
With the same operators, DAG planning matches ReAct in image quality with fewer cognitive-model requests, and typed diagnostics raise the first-repair success rate for each of the two tested cognitive models.

\paragraph{Limitations.}
\label{sec:limitations}
The open generator's rendering ability limits the gains on text-heavy AgentGen-Bench requests.
The checker validates only structural compatibility, so semantic and visual correctness depend on the cognitive and generation models.
Future work can extend the catalog with new operators and learn better planning strategies from the recorded plans and executions of earlier requests.

\section*{LLM Usage Statement}
{\parfillskip=0pt plus 0.25\linewidth OpenAI Codex and Anthropic Claude helped implement the agent framework, write experiment and analysis code, survey related work, draft and revise the manuscript, and prepare the LaTeX source and figure layouts.
Google Gemini helped polish the wording and grammar throughout the paper.
MiMo-V2.6-Flash and the image generators are components of the studied system, the benchmark judges (Sec.~\ref{sec:setup}) only score images, and all images in Figs.~\ref{fig:teaser}, \ref{fig:dag_cases}, and \ref{fig:qualitative} are generator outputs.
The authors reviewed all generated code, text, and figures and take full responsibility for every scientific claim made in this paper.\par}

\clearpage
\maketitlesupplementary
\suppressfloats[t]
\appendix
\setcounter{table}{0}\setcounter{figure}{0}
\renewcommand{\thetable}{A\arabic{table}}
\renewcommand{\thefigure}{A\arabic{figure}}
\renewcommand{\dblfloatpagefraction}{0.5}
\renewcommand{\floatpagefraction}{0.5}
\setcounter{dbltopnumber}{3}
\raggedbottom
\renewcommand{\dbltopfraction}{0.85}
\renewcommand{\dblfloatpagefraction}{0.9}
\makeatletter
\setlength{\@dblfptop}{0pt}\setlength{\@dblfpsep}{12pt}\setlength{\@dblfpbot}{0pt plus 1fil}
\setlength{\@fptop}{0pt}\setlength{\@fpsep}{8pt}\setlength{\@fpbot}{0pt plus 1fil}
\makeatother
\newcommand{\inputwide}[1]{\begingroup
  \renewenvironment{table}[1][t]{\begin{table*}[tbp]}{\end{table*}}%
  \input{#1}\endgroup}
\let\suppTable\table \let\endsuppTable\endtable
\renewenvironment{table}[1][t]{\suppTable[tp]}{\endsuppTable}
\newcommand{\inputtop}[1]{\begingroup\renewenvironment{table}[1][t]{\suppTable[t]}{\endsuppTable}\input{#1}\endgroup}
\makeatletter
\newcommand{\inputhere}[1]{\begingroup\renewenvironment{table}[1][t]{\par\ifdefined\nolinenumbers\nolinenumbers\fi\addvspace{\intextsep}\noindent\begin{minipage}{\linewidth}\def\@captype{table}\let\suppLabel\label\renewcommand{\label}[1]{\suppLabel{####1}\vspace{-5pt}}}{\end{minipage}\par\addvspace{\intextsep}\ifdefined\nolinenumbers\linenumbers\fi}\input{#1}\endgroup}
\makeatother
This supplementary material provides the details deferred from Secs.~\ref{sec:operator_catalog}, \ref{sec:plan_check}, \ref{sec:setup}, and \ref{sec:analysis}--\ref{sec:reliability} of the main paper (Secs.~\ref{app:catalog}--\ref{app:repair}) and more qualitative results (Sec.~\ref{app:qualitative}).

\section{Operator Catalog}
\label{app:catalog}
We build the operator catalog from 155,805 recorded trajectories, workflows, and plans collected from eight sources (Table~\ref{tab:supp_sources}).
Across these records, the twelve operators cover 86.68\% of all recorded actions and 89.11\% of the actions other than model routing.
Operators pass nine types of values along plan edges (Tables~\ref{tab:supp_types} and~\ref{tab:supp_operators}).

\begingroup\renewenvironment{table}[1][t]{\suppTable[t]}{\endsuppTable}\begin{table}[t]
\centering\footnotesize
\caption{\textbf{Sources of the operator catalog.} Coverage (Cov.) is the share of a source's actions that map to the 12 operators.}
\label{tab:supp_sources}
\setlength{\tabcolsep}{3pt}\renewcommand{\arraystretch}{1.1}
\begin{tabular}{@{}llrrr@{}}
\toprule
\textbf{Source} & \textbf{Type} & \textbf{Records} & \textbf{Actions} & \textbf{Cov.\ (\%)} \\
\midrule
OpenVisTool~\cite{xiang2026openvistool} & tool use & 42,048 & 296,993 & 96.31 \\
CanvasCraft SFT~\cite{zhu2026canvasagent} & tool use & 30,000 & 158,321 & 100.00 \\
CanvasCraft RL~\cite{zhu2026canvasagent} & tool sequences & 9,359 & 36,309 & 100.00 \\
Image styling~\cite{mukherjee2026styling} & action plans & 35,483 & 91,310 & 64.38 \\
Gen-Searcher~\cite{feng2026gensearcher} & search & 16,303 & 88,335 & 99.92 \\
ComfyGPT~\cite{huang2025comfygpt} & workflows & 12,571 & 287,821 & 68.10 \\
GenEvolve~\cite{chen2026genevolve} & tool use & 9,000 & 53,361 & 100.00 \\
Earlier planner & plans & 1,041 & 3,591 & 99.92 \\
\midrule
\textit{All} & & 155,805 & 1,016,041 & 86.68 \\
\bottomrule
\end{tabular}
\end{table}
\endgroup
\begingroup\renewenvironment{table}[1][t]{\suppTable[t]}{\endsuppTable}\begin{table}[t]
\centering\footnotesize
\caption{\textbf{Value types passed along plan edges.}}
\label{tab:supp_types}
\setlength{\tabcolsep}{4pt}\renewcommand{\arraystretch}{1.1}
\begin{tabular}{@{}l>{\raggedright\arraybackslash}p{0.68\linewidth}@{}}
\toprule
\textbf{Type} & \textbf{Content} \\
\midrule
\texttt{Query} & The original request text \\
\texttt{Intent} & Retrieval and reasoning targets extracted from the request \\
\texttt{Facts} & Textual evidence from search, reasoning, or skill notes \\
\texttt{RefImages} & Reference images with textual cues \\
\texttt{StyleSpec} & A selected visual style with prompt cues and optional style references \\
\texttt{ChecklistSpec} & Entities, verifiable constraints, and a checklist derived from the request \\
\texttt{CodeArtifact} & A rendered code artifact used as a layout reference \\
\texttt{PromptSpec} & A self-contained generation prompt \\
\texttt{Image} & A generated or edited image \\
\bottomrule
\end{tabular}
\end{table}
\endgroup
\begingroup
  \renewenvironment{table}[1][t]{\begin{table*}[tbp]}{\end{table*}}%
  \begin{table}[t]
\centering\footnotesize
\caption{\textbf{Operators.} Port types follow Table~\ref{tab:supp_types}. Brackets mark optional inputs, and \texttt{Edit} needs an image or reference images.}
\label{tab:supp_operators}
\setlength{\tabcolsep}{4pt}\renewcommand{\arraystretch}{1.12}
\begin{tabularx}{\linewidth}{@{}l>{\raggedright\arraybackslash}Xll@{}}
\toprule
\textbf{Operator} & \textbf{Function} & \textbf{Inputs} & \textbf{Output} \\
\midrule
\texttt{Analyze} & Extracts what to search for and what to reason about & Query & Intent \\
\texttt{Decompose} & Lists entities and verifiable visual constraints as a checklist & Query & ChecklistSpec \\
\texttt{SearchText} & Searches the web for textual facts & Query, [Intent] & Facts \\
\texttt{SearchImage} & Retrieves reference images of real entities & Query, [Intent] & RefImages \\
\texttt{SkillQuery} & Loads domain prompting notes from the skill bank & Query & Facts \\
\texttt{Reason} & Derives visual implications of implicit logical, temporal, spatial, or mathematical content & Query, [Intent], [Facts list] & Facts \\
\texttt{Rewrite} & Fuses the request and all acquired content into one generation prompt & Query, [Intent], [Facts list], [RefImages], [StyleSpec], [CodeArtifact] & PromptSpec \\
\texttt{SelectStyle} & Selects a visual pattern from a curated library & Query, [RefImages] & StyleSpec \\
\texttt{CodeExec} & Renders a code-backed layout artifact (SVG, HTML, or Three.js) & PromptSpec, [Intent] & CodeArtifact \\
\texttt{Generate} & Generates one image, optionally conditioned on references, style, or an artifact & PromptSpec, [RefImages], [StyleSpec], [CodeArtifact] & Image \\
\texttt{Edit} & Edits a generated image or an input image & PromptSpec, [Image], [RefImages], [StyleSpec] & Image \\
\texttt{RefineLoop} & Generates, checks the image against the checklist, and revises the prompt & PromptSpec, ChecklistSpec, [RefImages], [StyleSpec], [CodeArtifact] & Image \\
\bottomrule
\end{tabularx}
\end{table}
\endgroup

\section{Plan Checker}
\label{app:checks}
Before execution, the checker applies 17 checks, including the two planning rules (Table~\ref{tab:supp_checks}) and the parameter ranges in Table~\ref{tab:supp_params}.
All checks are deterministic and run locally, so checking a plan requires no model request.
Because the checker reports every violation that it finds in a plan at once, the agent can address several errors in a single repair.

\begingroup
  \renewenvironment{table}[1][t]{\begin{table*}[tbp]}{\end{table*}}%
  \begin{table}[t]
\centering\footnotesize
\caption{\textbf{Checks applied before execution.} Fourteen codes cover graph structure, types, parameters, size limits, and backend capabilities. \texttt{SEARCH\_UNAVAILABLE} checks the retrieval setup, and the last two codes enforce the planning rules of Sec.~\ref{sec:plan_check} in the main paper.}
\label{tab:supp_checks}
\setlength{\tabcolsep}{4pt}\renewcommand{\arraystretch}{1.12}
\begin{tabularx}{\linewidth}{@{}l>{\raggedright\arraybackslash}X@{}}
\toprule
\textbf{Code} & \textbf{Condition} \\
\midrule
\texttt{E\_SCHEMA} & The plan JSON does not parse into nodes, inputs, parameters, and an output. \\
\texttt{E\_DUP\_ID} & Two nodes share an identifier. \\
\texttt{E\_UNKNOWN\_OP} & A node uses an operator outside the catalog. \\
\texttt{E\_BAD\_REF} & An input references a missing node or an unavailable external input such as reference images. \\
\texttt{E\_CYCLE} & The graph contains a cycle. \\
\texttt{E\_UNKNOWN\_PORT} & A node binds a port that its operator does not declare. \\
\texttt{E\_MISSING\_PORT} & A required port is empty, or \texttt{Edit} receives neither an image nor references. \\
\texttt{E\_ARITY} & A list-valued port receives a single source, or a single-valued port receives a list. \\
\texttt{E\_TYPE} & A source produces a type that the receiving port does not accept. Prompt ports also accept the request. \\
\texttt{E\_OUTPUT} & The designated output is missing or does not produce an image. \\
\texttt{E\_ORPHAN} & A node has no path to the output. \\
\texttt{E\_BUDGET} & The plan exceeds 15 nodes or an operator limit: one \texttt{Generate}, \texttt{RefineLoop}, \texttt{CodeExec}, \texttt{SelectStyle}, and \texttt{SearchImage}, and two \texttt{SearchText} and \texttt{Edit}. \\
\texttt{E\_PARAM} & A parameter is undeclared or out of range (Table~\ref{tab:supp_params}). \\
\texttt{E\_CAPABILITY} & A node needs a capability that the configuration lacks: image-conditioned generation for references, edits, or artifacts, or a code-rendering backend for \texttt{CodeExec}. \\
\texttt{SEARCH\_UNAVAILABLE} & The plan uses \texttt{SearchText} or \texttt{SearchImage} without a configured retrieval backend. \\
\texttt{GATE\_SEARCH\_REQUIRED} & The request is labeled incomplete or ambiguous, retrieval is available, and the plan has no \texttt{SearchText}. \\
\texttt{BARE\_GENERATE\_FORBIDDEN} & \texttt{Generate} reads its prompt directly from the request instead of a \texttt{Rewrite} node. \\
\bottomrule
\end{tabularx}
\end{table}
\endgroup
\begin{table}[t]
\centering\footnotesize
\caption{\textbf{Operator parameters accepted by the checker.} Operators not listed in the table take no parameters.}
\label{tab:supp_params}
\setlength{\tabcolsep}{3pt}\renewcommand{\arraystretch}{1.1}
\begin{tabular}{@{}l>{\raggedright\arraybackslash}p{0.72\linewidth}@{}}
\toprule
\textbf{Operator} & \textbf{Parameters} \\
\midrule
\texttt{SearchText} & \texttt{top\_k} $\in[1,5]$ (default 3), \texttt{search\_query} (string) \\
\texttt{SearchImage} & \texttt{top\_k} $\in[1,5]$ (default 2), \texttt{role} $\in\{$identity, style, layout$\}$ \\
\texttt{SelectStyle} & \texttt{source} $\in\{$auto, pattern, refs$\}$ \\
\texttt{CodeExec} & \texttt{code\_type} $\in\{$auto, svg, html\_css, threejs$\}$ \\
\texttt{Edit} & \texttt{input\_fidelity} $\in\{$low, high$\}$ (default high) \\
\texttt{RefineLoop} & \texttt{max\_iter} $\in[0,2]$ (default 2) \\
\bottomrule
\end{tabular}
\end{table}

\section{Baselines and Evaluation}
\label{app:baselines}
We adapt the official implementations of GEMS~\cite{he2026gems}, Mind-Brush~\cite{he2026mindbrush_sa}, GenRouter~\cite{chen2026genrouter}, and SCOPE~\cite{ren2026scope} to the same backends.
Each agent uses MiMo-V2.6-Flash as its cognitive model, Qwen-Image-2.1 for generation and editing, and the frozen retrieval environment for search.
All agents keep their original workflows and iteration limits, such as five refinement iterations in GEMS and up to three generations in SCOPE.
GEMS and GenRouter run with text-only input, so their Mind-Bench scores cover the 350 T2I requests.

We build the frozen retrieval environment from the retrieval caches that earlier runs of the compared methods produced on the same benchmarks.
It contains 77,626 queries, 124,648 text documents, and 54,314 reference images.
Uncached queries search this corpus with BM25~\cite{robertson2009bm25} and Qwen3-Embedding-0.6B~\cite{zhang2025qwen3embedding}.
Each document provides a title and a snippet, and 39,055 documents (31\%) also store the page body, which only Mind-Brush reads.

In Table~\ref{tab:main} of the main paper, all Qwen-Image-2.1 methods generate and edit at $1024^2$, and a failed execution restarts for up to six attempts.
On T2I-CoReBench, open generators use seeds 0--3, and commercial APIs receive four requests.
On the other benchmarks, open generators use a fixed seed.
Across all benchmarks, Nano Banana Pro returns no image in 135 executions, and GPT-Image-2 declines 17 requests for safety reasons.
For Table~\ref{tab:backend_generalization} of the main paper, the Qwen-Image-2512/Edit-2511 group uses 50 generation steps, 40 editing steps, and a guidance scale of 4.
The GPT-Image-2 group uses $1024^2$ at medium quality and also gives the \method{} w/ GPT-Image-2 results in Table~\ref{tab:main}.

\section{Plan Statistics}
\label{app:plan_stats}
We compute plan statistics for every benchmark on the last executed plan of each successful execution (Table~\ref{tab:plan_statistics}).
Plans for AgentGen-Bench and Mind-Bench T2I contain the most operators, 5.58 and 5.54 on average, and they also call \texttt{SearchImage} for reference images most often.

\begin{table}[t]
\centering\footnotesize
\caption{\textbf{Plans synthesized for different benchmarks.} Statistics cover 7,368 successful GenDAG executions with Qwen-Image-2.1 in Table~\ref{tab:main} of the main paper. Nodes reports the mean operator count, and Graphs counts distinct structures up to node renaming, with port bindings retained and parameters ignored. The All row counts a structure shared by several benchmarks once. Text, Image, Reason, and Refine give percentages of plans containing \texttt{SearchText}, \texttt{SearchImage}, \texttt{Reason}, and \texttt{RefineLoop}, which consumes a \texttt{Decompose} checklist.}
\label{tab:plan_statistics}
\setlength{\tabcolsep}{2.6pt}\renewcommand{\arraystretch}{1.12}
\begin{tabular}{lrrrrrr}
\toprule
\multirow{2}{*}[-\dimexpr(\aboverulesep+\belowrulesep+\cmidrulewidth)/2\relax]{\textbf{Benchmark}} & \multirow{2}{*}[-\dimexpr(\aboverulesep+\belowrulesep+\cmidrulewidth)/2\relax]{\textbf{Nodes}} & \multirow{2}{*}[-\dimexpr(\aboverulesep+\belowrulesep+\cmidrulewidth)/2\relax]{\textbf{Graphs}} & \multicolumn{4}{c}{\textbf{Plans containing (\%)}} \\
\cmidrule(l){4-7}
 & & & \textbf{Text} & \textbf{Image} & \textbf{Reason} & \textbf{Refine} \\
\midrule
WISE & 4.77 & 26 & 81.50 & 9.00 & 55.80 & 2.80 \\
GenEval2 & 4.04 & 19 & 9.25 & 0.00 & 17.88 & 74.13 \\
CoReBench comp. & 4.50 & 31 & 15.21 & 0.42 & 34.03 & 91.04 \\
CoReBench reas. & 4.74 & 50 & 27.99 & 0.21 & 70.14 & 63.40 \\
AgentGen & 5.58 & 59 & 81.95 & 26.07 & 33.82 & 58.82 \\
Mind-Bench T2I & 5.54 & 29 & 96.00 & 52.29 & 43.71 & 12.86 \\
Mind-Bench I2I & 5.27 & 37 & 86.00 & 10.67 & 69.33 & 35.33 \\
\midrule
\textit{All} & 4.75 & 136 & 40.61 & 6.73 & 50.50 & 58.31 \\
\bottomrule
\end{tabular}
\end{table}

\section{DAG Planning versus ReAct}
\label{app:react}
The ReAct agent~\cite{yao2022react} chooses one operator per step and checks each step before it runs, with at most 15 nodes per round and one follow-up round.
After two failed repairs of an action, it falls back to the simplest valid actions, whereas \method{} falls back to the closest example plan.
Including all attempts on WISE and T2I-CoReBench, ReAct uses 66.26 and 87.07 cognitive-model requests per execution, compared with 54.80 and 71.32 for \method{}.
ReAct also issues 1.66 and 0.43 retrieval requests, versus 1.32 and 0.29 for \method{}.
Both agents issue similar numbers of generation and editing requests on these benchmarks, with 3.28 and 5.02 for \method{} and 3.16 and 4.86 for ReAct.

\section{Repair Cohorts}
\label{app:repair}
The repair experiment in Table~\ref{tab:reliability} of the main paper uses two cohorts of invalid first plans.
The MiMo-V2.5 cohort contains all 42 invalid first plans from 2,080 planning-only sessions.
The MiMo-V2.6-Flash cohort contains all 48 invalid first plans (0.90\%) from the 5,320 main-system executions.
In both cohorts, the most common error is an undeclared parameter, which appears in 21 of the 42 plans and 30 of the 48.
The second most common error is a reference to a nonexistent node, found in 10 and 6 plans (Table~\ref{tab:supp_checks}).

\section{More Qualitative Results}
\label{app:qualitative}
All examples show the first execution of each request, and badges give benchmark scores as in Fig.~\ref{fig:qualitative} of the main paper.

\begin{figure*}[t]
\centering
\includegraphics[width=\linewidth]{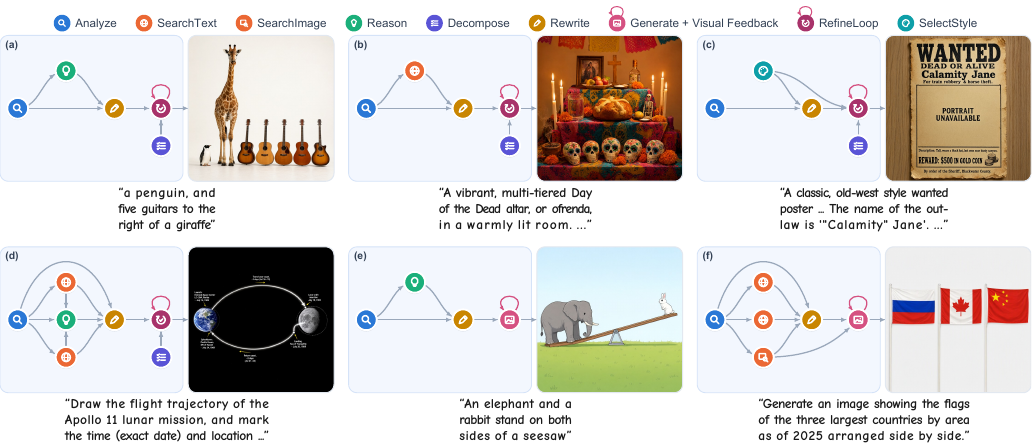}
\caption{\textbf{More query-adaptive plans of \method{} with Qwen-Image-2.1.} Each panel shows the executed plan, the request, and the final image in the notation of Fig.~\ref{fig:dag_cases} of the main paper. The six requests come from GenEval2 (a), T2I-CoReBench (b, c), AgentGen-Bench (d), WISE (e), and Mind-Bench (f). Long requests are shortened, and edges from the request are omitted.}
\label{fig:supp_plans}
\end{figure*}

\paragraph{Query-Adaptive Plans.} Six more plans show graph structures absent from Fig.~\ref{fig:dag_cases} of the main paper (Fig.~\ref{fig:supp_plans}).

\begin{figure*}[t]
\centering
\includegraphics[width=\linewidth]{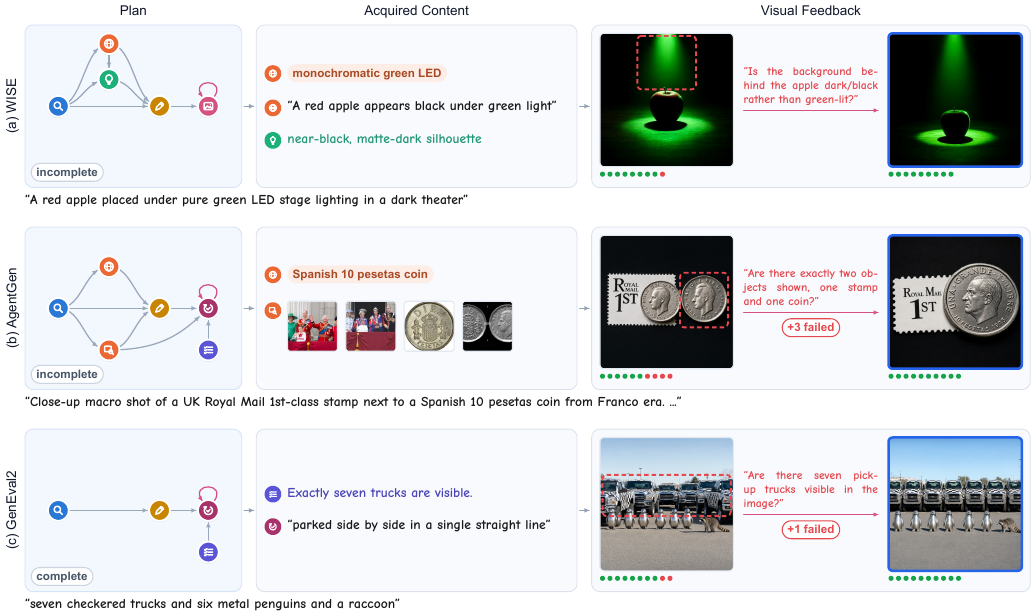}
\caption{\textbf{Case studies of \method{} with Qwen-Image-2.1.} Each row shows the executed plan, the content that the plan acquires or derives, and the visual feedback. The feedback panel shows the plan output with its post-execution checks (green pass, red fail) and a dashed box around the failed region. It also shows one failed question with the number of further failures and the regenerated image in the blue frame. Icons follow Fig.~\ref{fig:supp_plans}, and long requests under each row are shortened. Requests (b) and (c) reappear in Figs.~\ref{fig:supp_curated_b} and~\ref{fig:supp_curated_a}.}
\label{fig:supp_cases}
\end{figure*}

\paragraph{Case Studies.} Three case studies from WISE, AgentGen-Bench, and GenEval2 follow requests through planning, acquisition, generation, and visual feedback (Fig.~\ref{fig:supp_cases}).

\begin{figure*}[t]
\centering
\includegraphics[width=\linewidth]{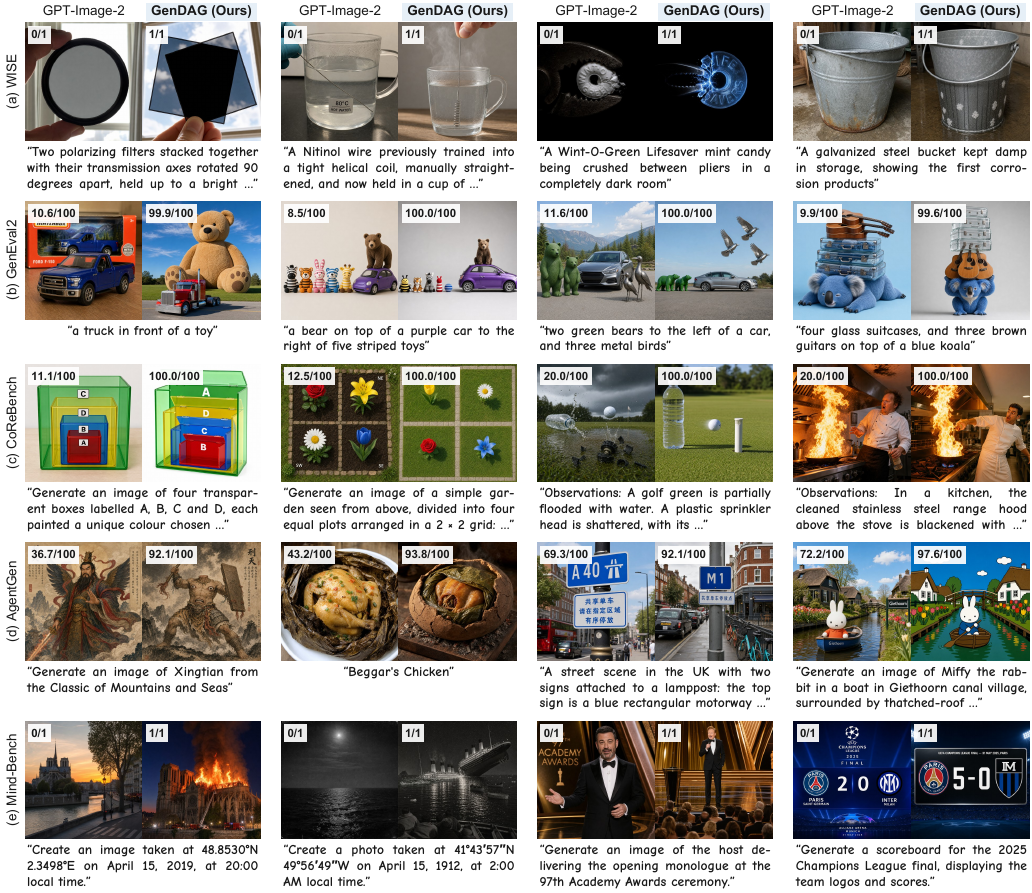}
\caption{\textbf{Direct GPT-Image-2 generation (left of each pair) versus \method{} with GPT-Image-2 (right).}}
\label{fig:supp_gpt}
\end{figure*}

\paragraph{Agentic Generation with GPT-Image-2.} With GPT-Image-2 as the generator, \method{} passes 115 WISE requests that direct generation fails, and it passes 67 more WISE requests in total.
Twenty example pairs from all five benchmarks are shown side by side in Fig.~\ref{fig:supp_gpt}.

\begin{figure*}[t]
\centering
\includegraphics[width=\linewidth]{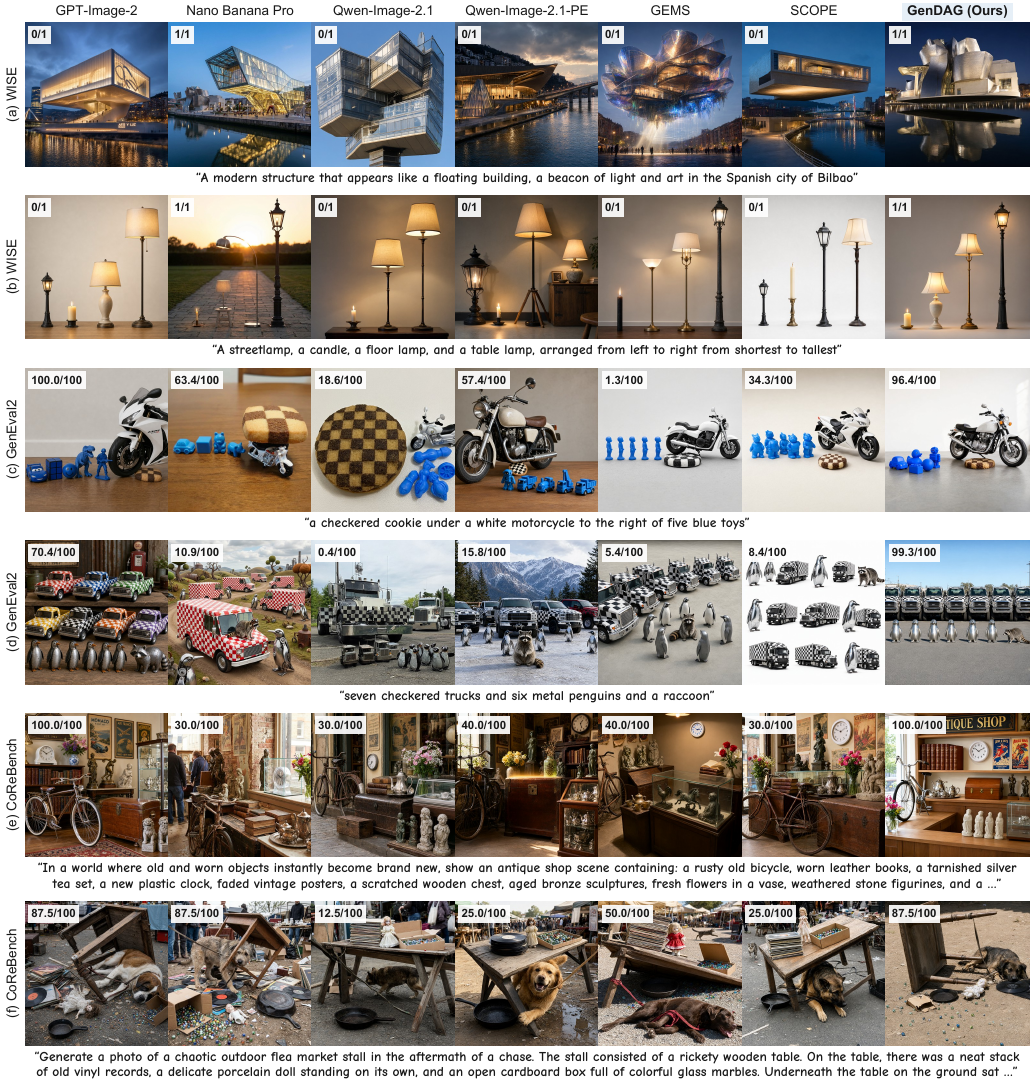}
\caption{\textbf{More qualitative results on WISE, GenEval2, and T2I-CoReBench.}}
\label{fig:supp_curated_a}
\end{figure*}

\begin{figure*}[t]
\centering
\includegraphics[width=\linewidth]{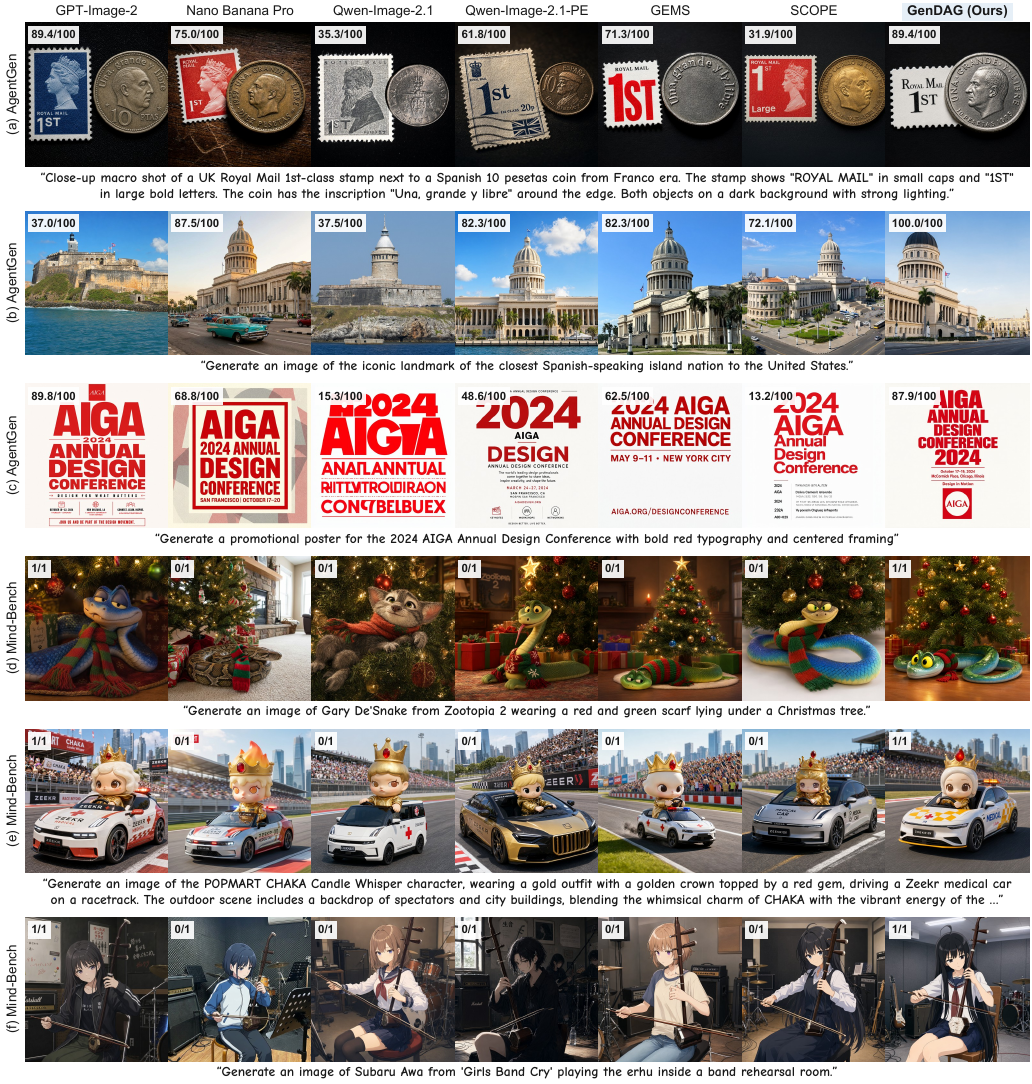}
\caption{\textbf{More qualitative results on AgentGen-Bench and Mind-Bench.}}
\label{fig:supp_curated_b}
\end{figure*}

\paragraph{More Comparisons.} Additional qualitative results on all five benchmarks are shown in Figs.~\ref{fig:supp_curated_a} and~\ref{fig:supp_curated_b}.

\FloatBarrier
\clearpage
\renewcommand{\dblfloatpagefraction}{0.5}

{\small\begin{thebibliography}{40}
\providecommand{\natexlab}[1]{#1}
\providecommand{\url}[1]{\texttt{#1}}
\expandafter\ifx\csname urlstyle\endcsname\relax
  \providecommand{\doi}[1]{doi: #1}\else
  \providecommand{\doi}{doi: \begingroup \urlstyle{rm}\Url}\fi

\bibitem[Bai et~al.(2025)Bai, Cai, Chen, et~al.]{bai2025qwen3vl}
Shuai Bai, Yuxuan Cai, Ruizhe Chen, et~al.
\newblock {Qwen3-VL} technical report.
\newblock \emph{arXiv preprint arXiv:2511.21631}, 2025.

\bibitem[Chen et~al.(2025)Chen, Shi, Zhang, and Shi]{chen2025t2icopilot}
Chieh-Yun Chen, Min Shi, Gong Zhang, and Humphrey Shi.
\newblock {T2I-Copilot}: A training-free multi-agent text-to-image system for enhanced prompt interpretation and interactive generation.
\newblock In \emph{Proceedings of the IEEE/CVF International Conference on Computer Vision}, 2025.

\bibitem[Chen et~al.(2026{\natexlab{a}})Chen, Hou, Shu, Ruan, Xu, Guo, and Chen]{chen2026genrouter}
Harold~Haodong Chen, Zhiyu Hou, Wen-Jie Shu, Weilin Ruan, Yingjie Xu, Litao Guo, and Ying-Cong Chen.
\newblock {GenRouter}: Unified workflow routing for agentic image generation.
\newblock \emph{arXiv preprint arXiv:2608.16721}, 2026{\natexlab{a}}.

\bibitem[Chen et~al.(2026{\natexlab{b}})Chen, Xing, Ye, Geng, Lin, Lai, He, Zhai, Gao, and Zhu]{chen2026genevolve}
Sixiang Chen, Zhaohu Xing, Tian Ye, Xinyu Geng, Yunlong Lin, Jianyu Lai, Xuanhua He, Fuxiang Zhai, Jialin Gao, and Lei Zhu.
\newblock {GenEvolve}: Self-evolving image generation agents via tool-orchestrated visual experience distillation.
\newblock In \emph{Advances in Neural Information Processing Systems}, 2026{\natexlab{b}}.

\bibitem[Chen et~al.(2026{\natexlab{c}})Chen, Yu, Tian, Song, Liang, Jia, Cheng, Li, Yuan, Wang, Wu, Lai, and Yue]{chen2026memogen}
Wenshuo Chen, Kuimou Yu, Bowen Tian, Jianfei Song, Shaofeng Liang, Haozhe Jia, Kan Cheng, Haosen Li, Kaishen Yuan, Lei Wang, Jiemin Wu, Songning Lai, and Yutao Yue.
\newblock {MemoGen}: Can past experience improve future text-to-image generation?
\newblock \emph{arXiv preprint arXiv:2606.03243}, 2026{\natexlab{c}}.

\bibitem[Esser et~al.(2024)Esser, Kulal, Blattmann, Entezari, M{\"u}ller, Saini, Levi, Lorenz, Sauer, Boesel, Podell, Dockhorn, English, Lacey, Goodwin, Marek, and Rombach]{esser2024sd3}
Patrick Esser, Sumith Kulal, Andreas Blattmann, Rahim Entezari, Jonas M{\"u}ller, Harry Saini, Yam Levi, Dominik Lorenz, Axel Sauer, Frederic Boesel, Dustin Podell, Tim Dockhorn, Zion English, Kyle Lacey, Alex Goodwin, Yannik Marek, and Robin Rombach.
\newblock Scaling rectified flow transformers for high-resolution image synthesis.
\newblock In \emph{Proceedings of the 41st International Conference on Machine Learning}, 2024.

\bibitem[Fang et~al.(2025)Fang, Duan, Wang, Huang, Li, Yan, Tian, Zeng, Zhao, Dai, Liu, and Li]{fang2025got}
Rongyao Fang, Chengqi Duan, Kun Wang, Linjiang Huang, Hao Li, Shilin Yan, Hao Tian, Xingyu Zeng, Rui Zhao, Jifeng Dai, Xihui Liu, and Hongsheng Li.
\newblock {GoT}: Unleashing reasoning capability of multimodal large language model for visual generation and editing.
\newblock In \emph{Advances in Neural Information Processing Systems}, 2025.

\bibitem[Feng et~al.(2026)Feng, Zhang, Chen, Lin, Fan, Jiang, Li, Zheng, Wang, and Yue]{feng2026gensearcher}
Kaituo Feng, Manyuan Zhang, Shuang Chen, Yunlong Lin, Kaixuan Fan, Yilei Jiang, Hongyu Li, Dian Zheng, Chenyang Wang, and Xiangyu Yue.
\newblock {Gen-Searcher}: Reinforcing agentic search for image generation.
\newblock In \emph{Advances in Neural Information Processing Systems}, 2026.

\bibitem[Gupta and Kembhavi(2023)]{gupta2022visprog}
Tanmay Gupta and Aniruddha Kembhavi.
\newblock Visual programming: Compositional visual reasoning without training.
\newblock In \emph{Proceedings of the IEEE/CVF Conference on Computer Vision and Pattern Recognition}, 2023.

\bibitem[He et~al.(2026{\natexlab{a}})He, Ye, Huang, Jiang, Zhang, Zhu, Zhang, Zhang, and Li]{he2026mindbrush_sa}
Jun He, Junyan Ye, Zilong Huang, Dongzhi Jiang, Chenjue Zhang, Leqi Zhu, Renrui Zhang, Xiang Zhang, and Weijia Li.
\newblock {Mind-Brush}: Integrating agentic cognitive search and reasoning into image generation.
\newblock In \emph{Proceedings of the SIGGRAPH Asia 2026 Conference Papers}, 2026{\natexlab{a}}.

\bibitem[He et~al.(2026{\natexlab{b}})He, Huang, Qu, Li, Zhu, Cheng, and Yang]{he2026gems}
Zefeng He, Siyuan Huang, Xiaoye Qu, Yafu Li, Tong Zhu, Yu Cheng, and Yang Yang.
\newblock {GEMS}: Agent-native multimodal generation with memory and skills.
\newblock \emph{arXiv preprint arXiv:2603.28088}, 2026{\natexlab{b}}.

\bibitem[Hu et~al.(2023)Hu, Liu, Kasai, Wang, Ostendorf, Krishna, and Smith]{hu2023tifa}
Yushi Hu, Benlin Liu, Jungo Kasai, Yizhong Wang, Mari Ostendorf, Ranjay Krishna, and Noah~A Smith.
\newblock {TIFA}: Accurate and interpretable text-to-image faithfulness evaluation with question answering.
\newblock In \emph{Proceedings of the IEEE/CVF International Conference on Computer Vision}, 2023.

\bibitem[Huang et~al.(2025)Huang, Ma, Zhao, Wu, Ji, Zhang, Hu, Sun, and Ji]{huang2025comfygpt}
Oucheng Huang, Yuhang Ma, Zeng Zhao, Mingrui Wu, Jiayi Ji, Rongsheng Zhang, Zhipeng Hu, Xiaoshuai Sun, and Rongrong Ji.
\newblock {ComfyGPT}: A self-optimizing multi-agent system for comprehensive {ComfyUI} workflow generation.
\newblock \emph{arXiv preprint arXiv:2503.17671}, 2025.

\bibitem[Jiang et~al.(2025)Jiang, Guo, Zhang, Zong, Li, Zhuo, Yan, Heng, and Li]{jiang2025t2ir1}
Dongzhi Jiang, Ziyu Guo, Renrui Zhang, Zhuofan Zong, Hao Li, Le Zhuo, Shilin Yan, Pheng-Ann Heng, and Hongsheng Li.
\newblock {T2I-R1}: Reinforcing image generation with collaborative semantic-level and token-level {CoT}.
\newblock In \emph{Advances in Neural Information Processing Systems}, 2025.

\bibitem[Kamath et~al.(2025)Kamath, Chang, Krishna, Zettlemoyer, Hu, and Ghazvininejad]{kamath2025geneval2}
Amita Kamath, Kai-Wei Chang, Ranjay Krishna, Luke Zettlemoyer, Yushi Hu, and Marjan Ghazvininejad.
\newblock {GenEval 2}: Addressing benchmark drift in text-to-image evaluation.
\newblock \emph{arXiv preprint arXiv:2512.16853}, 2025.

\bibitem[Kim et~al.(2024)Kim, Moon, Tabrizi, Lee, Mahoney, Keutzer, and Gholami]{kim2023llmcompiler}
Sehoon Kim, Suhong Moon, Ryan Tabrizi, Nicholas Lee, Michael~W. Mahoney, Kurt Keutzer, and Amir Gholami.
\newblock An {LLM} compiler for parallel function calling.
\newblock In \emph{Proceedings of the 41st International Conference on Machine Learning}, 2024.

\bibitem[Li et~al.(2026{\natexlab{a}})Li, Wang, Hu, Huang, Chen, Ou, Tao, Wan, Qi, and Feng]{li2025core}
Ouxiang Li, Yuan Wang, Xinting Hu, Huijuan Huang, Rui Chen, Jiarong Ou, Xin Tao, Pengfei Wan, Xiaojuan Qi, and Fuli Feng.
\newblock Easier painting than thinking: Can text-to-image models set the stage, but not direct the play?
\newblock In \emph{International Conference on Learning Representations}, 2026{\natexlab{a}}.

\bibitem[Li et~al.(2026{\natexlab{b}})Li, Liu, Liu, Zhou, Wu, Chen, Xie, Wu, and Sun]{li2026comfyclaw}
Zongxia Li, Dawei Liu, Fuxiao Liu, Yuhang Zhou, Xiyang Wu, Jingxi Chen, Jing Xie, Xiaomin Wu, and Lichao Sun.
\newblock {ComfyClaw}: Self-evolving skill harnesses for image generation workflows.
\newblock \emph{arXiv preprint arXiv:2607.01709}, 2026{\natexlab{b}}.

\bibitem[Mukherjee et~al.(2026)Mukherjee, Petrangeli, Kveton, Bui, Dernoncourt, and Mukherjee]{mukherjee2026styling}
Subhojyoti Mukherjee, Stefano Petrangeli, Branislav Kveton, Trung Bui, Franck Dernoncourt, and Arko Mukherjee.
\newblock Agentic planning with reasoning for image styling via offline {RL}.
\newblock \emph{arXiv preprint arXiv:2603.07148}, 2026.

\bibitem[Niu et~al.(2026)Niu, Ning, Zheng, Jin, Lin, Jin, Liao, Feng, Meng, Ning, Zhu, and Yuan]{niu2026wise}
Yuwei Niu, Munan Ning, Mengren Zheng, Weiyang Jin, Bin Lin, Peng Jin, Jiaqi Liao, Chaoran Feng, Fanqing Meng, Kunpeng Ning, Bin Zhu, and Li Yuan.
\newblock {WISE}: World knowledge-informed semantic evaluation for text-to-image generation.
\newblock In \emph{Proceedings of the 43rd International Conference on Machine Learning}, 2026.

\bibitem[Podell et~al.(2024)Podell, English, Lacey, Blattmann, Dockhorn, M{\"u}ller, Penna, and Rombach]{podell2023sdxl}
Dustin Podell, Zion English, Kyle Lacey, Andreas Blattmann, Tim Dockhorn, Jonas M{\"u}ller, Joe Penna, and Robin Rombach.
\newblock {SDXL}: Improving latent diffusion models for high-resolution image synthesis.
\newblock In \emph{International Conference on Learning Representations}, 2024.

\bibitem[{Qwen Team}(2026)]{teamqwen}
{Qwen Team}.
\newblock {Qwen-Image-2.1}: Compact, efficient, and unified image creation.
\newblock Qwen Blog, 2026.

\bibitem[Ren et~al.(2026)Ren, Yan, Zhao, Fang, Zeng, Zhang, Xu, Ma, Huang, Xu, Huang, Wang, Chen, Chen, Huang, and Zhao]{ren2026scope}
Tianfei Ren, Zhipeng Yan, Yiming Zhao, Zhen Fang, Yu Zeng, Guohui Zhang, Hang Xu, Xiaoxiao Ma, Shiting Huang, Ke Xu, Wenxuan Huang, Lionel~Z. Wang, Lin Chen, Zehui Chen, Jie Huang, and Feng Zhao.
\newblock {SCOPE}: Structured decomposition and conditional skill orchestration for complex image generation.
\newblock \emph{arXiv preprint arXiv:2605.08043}, 2026.

\bibitem[Robertson and Zaragoza(2009)]{robertson2009bm25}
Stephen Robertson and Hugo Zaragoza.
\newblock The probabilistic relevance framework: {BM25} and beyond.
\newblock \emph{Foundations and Trends in Information Retrieval}, 2009.

\bibitem[Rombach et~al.(2022)Rombach, Blattmann, Lorenz, Esser, and Ommer]{rombach2022ldm}
Robin Rombach, Andreas Blattmann, Dominik Lorenz, Patrick Esser, and Bj{\"o}rn Ommer.
\newblock High-resolution image synthesis with latent diffusion models.
\newblock In \emph{Proceedings of the IEEE/CVF Conference on Computer Vision and Pattern Recognition}, 2022.

\bibitem[Shinn et~al.(2023)Shinn, Cassano, Berman, Gopinath, Narasimhan, and Yao]{shinn2023reflexion}
Noah Shinn, Federico Cassano, Edward Berman, Ashwin Gopinath, Karthik Narasimhan, and Shunyu Yao.
\newblock {Reflexion}: Language agents with verbal reinforcement learning.
\newblock In \emph{Advances in Neural Information Processing Systems}, 2023.

\bibitem[Wang et~al.(2026)Wang, Feng, Yu, Liu, Nie, Lin, Liu, Huang, Lin, Chen, and Wei]{wang2026searchgen}
Haozhe Wang, Weijia Feng, Jinpeng Yu, Che Liu, Ping Nie, Fangzhen Lin, Jiaming Liu, Ruihua Huang, Jimmy Lin, Wenhu Chen, and Cong Wei.
\newblock Search beyond what can be taught: Evolving the knowledge boundary in agentic visual generation.
\newblock \emph{arXiv preprint arXiv:2607.05382}, 2026.

\bibitem[Wang et~al.(2024)Wang, Li, Li, and Liu]{wang2024genartist}
Zhenyu Wang, Aoxue Li, Zhenguo Li, and Xihui Liu.
\newblock {GenArtist}: Multimodal {LLM} as an agent for unified image generation and editing.
\newblock In \emph{Advances in Neural Information Processing Systems}, 2024.

\bibitem[Wu et~al.(2025)Wu, Li, Zhou, et~al.]{wu2025qwenimage}
Chenfei Wu, Jiahao Li, Jingren Zhou, et~al.
\newblock {Qwen-Image} technical report.
\newblock \emph{arXiv preprint arXiv:2508.02324}, 2025.

\bibitem[Xia et~al.(2026)Xia, Hu, Sun, Xu, Xu, Wang, Xu, and Jiang]{xia2026grasp}
Tianle Xia, Lingxiang Hu, Yiding Sun, Ming Xu, Lan Xu, Siying Wang, Wei Xu, and Jie Jiang.
\newblock {GraSP}: Graph-structured skill compositions for {LLM} agents.
\newblock \emph{arXiv preprint arXiv:2604.17870}, 2026.

\bibitem[Xiang et~al.(2026)Xiang, Zhang, Ma, Cheng, Ma, Feng, Zhang, Wang, Wu, Dai, and Lu]{xiang2026openvistool}
Changhao Xiang, Shilin Zhang, Zheng Ma, Kanzhi Cheng, Ruize Ma, Yi Feng, Jianbing Zhang, Zhi Wang, Zhen Wu, Xinyu Dai, and Lewei Lu.
\newblock {OpenVisTool}: An open recipe for synthesizing instructive visual tool-use trajectories.
\newblock \emph{arXiv preprint arXiv:2608.08557}, 2026.

\bibitem[{Xiaomi LLM-Core Team}(2026)]{xiaomi2026mimov26}
{Xiaomi LLM-Core Team}.
\newblock {MiMo-V2.6}: Scaling reinforcement learning towards self-improvement.
\newblock \emph{arXiv preprint arXiv:2610.11959}, 2026.

\bibitem[Yang et~al.(2026)Yang, Guo, Fan, Cheng, and Liu]{yang2026m3}
Bangji Yang, Ruihan Guo, Jiajun Fan, Chaoran Cheng, and Ge Liu.
\newblock {M3}: High-fidelity text-to-image generation via multi-modal, multi-agent and multi-round visual reasoning.
\newblock \emph{arXiv preprint arXiv:2602.06166}, 2026.

\bibitem[Yang et~al.(2024)Yang, Wang, Li, Lin, Lin, Liu, and Wang]{yang2024idea2img}
Zhengyuan Yang, Jianfeng Wang, Linjie Li, Kevin Lin, Chung-Ching Lin, Zicheng Liu, and Lijuan Wang.
\newblock {Idea2Img}: Iterative self-refinement with {GPT-4V(ision)} for automatic image design and generation.
\newblock In \emph{European Conference on Computer Vision}, 2024.

\bibitem[Yao et~al.(2023)Yao, Zhao, Yu, Du, Shafran, Narasimhan, and Cao]{yao2022react}
Shunyu Yao, Jeffrey Zhao, Dian Yu, Nan Du, Izhak Shafran, Karthik Narasimhan, and Yuan Cao.
\newblock {ReAct}: Synergizing reasoning and acting in language models.
\newblock In \emph{International Conference on Learning Representations}, 2023.

\bibitem[Ye et~al.(2026)Ye, He, Huang, Jiang, Yang, Chen, and Li]{ye2026genclaw}
Junyan Ye, Jun He, Zilong Huang, Dongzhi Jiang, Xuan Yang, Rui Chen, and Weijia Li.
\newblock {GenClaw}: Code-driven agentic image generation.
\newblock \emph{arXiv preprint arXiv:2605.30248}, 2026.

\bibitem[Zhang et~al.(2025)Zhang, Li, Long, Zhang, Lin, Yang, Xie, Yang, Liu, Lin, Huang, and Zhou]{zhang2025qwen3embedding}
Yanzhao Zhang, Mingxin Li, Dingkun Long, Xin Zhang, Huan Lin, Baosong Yang, Pengjun Xie, An Yang, Dayiheng Liu, Junyang Lin, Fei Huang, and Jingren Zhou.
\newblock {Qwen3 Embedding}: Advancing text embedding and reranking through foundation models.
\newblock \emph{arXiv preprint arXiv:2506.05176}, 2025.

\bibitem[Zhang et~al.(2026)Zhang, Li, Zhang, Gao, Yan, Jiang, Tang, Yin, Wu, Chen, Xu, Shu, Zhang, Xu, Chen, Wang, Liu, Zhou, Zhang, Zhao, and Wu]{zhang2026qia}
Zekai Zhang, Jiahao Li, Jie Zhang, Kaiyuan Gao, Kun Yan, Lihan Jiang, Ningyuan Tang, Shengming Yin, Tianhe Wu, Xiaoyue Chen, Xiao Xu, Yan Shu, Yanran Zhang, Yixian Xu, Yuxiang Chen, Zhendong Wang, Zihao Liu, Zikai Zhou, Huishuai Zhang, Dongyan Zhao, and Chenfei Wu.
\newblock {Qwen-Image-Agent}: Bridging the context gap in real-world image generation.
\newblock \emph{arXiv preprint arXiv:2606.26907}, 2026.

\bibitem[Zhou and Yang(2026)]{zhou2026stride}
Yan Zhou and Yang Yang.
\newblock {Stride}: Scale-aware multi-agent evolutionary refinement for infographic generation.
\newblock \emph{arXiv preprint arXiv:2610.13502}, 2026.

\bibitem[Zhu et~al.(2026)Zhu, Yang, Weng, Lu, Yao, Ye, Ma, and Jiang]{zhu2026canvasagent}
Hairui Zhu, Yiying Yang, Tengjin Weng, Ziyu Lu, Xiao Yao, Xiaoyang Ye, Lin Ma, and Wenhao Jiang.
\newblock {CanvasAgent}: Enabling complex image creation and editing via visual tool orchestration.
\newblock \emph{arXiv preprint arXiv:2607.05465}, 2026.

\end{thebibliography}
}
\end{document}